%% file: main.tex
\documentclass{article}

\usepackage{microtype}
\usepackage{graphicx}
\usepackage{subfigure}
\usepackage{booktabs} 
\usepackage{tcolorbox}
\usepackage{diagbox}
\usepackage{multirow}
\usepackage{multicol}
\usepackage{hyperref}

\usepackage[accepted]{icml2024}

\usepackage{amsmath}
\usepackage{amssymb}
\usepackage{mathtools}
\usepackage{amsthm}

\usepackage{bm}

\input{notations-bold-bm}

\usepackage[capitalize,noabbrev]{cleveref}

\theoremstyle{plain}
\newtheorem{theorem}{Theorem}

\theoremstyle{definition}
\newtheorem{definition}{Definition}

\theoremstyle{remark}

\usepackage[textsize=tiny]{todonotes}

\icmltitlerunning{NPSVC++: Nonparallel Classifiers Encounter Representation Learning}

\begin{document}

\twocolumn[
\icmltitle{NPSVC++: Nonparallel Classifiers Encounter Representation Learning}



\icmlsetsymbol{equal}{*}

\begin{icmlauthorlist}
\icmlauthor{Junhong Zhang}{CSSE}
\icmlauthor{Zhihui Lai}{CSSE}
\icmlauthor{Jie Zhou}{CSSE}
\icmlauthor{Guangfei Liang}{CSSE}
\end{icmlauthorlist}

\icmlaffiliation{CSSE}{College of Computer Science and Software Engineering, Shenzhen University, Shenzhen, China}

\icmlcorrespondingauthor{Zhihui Lai}{lai\_zhi\_hui@163.com}

\icmlkeywords{Machine Learning, ICML}

\vskip 0.3in
]



\printAffiliationsAndNotice{}  

\begin{abstract}
This paper focuses on a specific family of classifiers called nonparallel support vector classifiers (NPSVCs).
Different from typical classifiers, the training of an NPSVC involves the minimization of multiple objectives,
resulting in the potential concerns of \emph{feature suboptimality} and \emph{class dependency}.
Consequently, no effective learning scheme has been established to improve NPSVCs' performance through  representation learning, especially deep learning.
To break this bottleneck, we develop NPSVC++ based on multi-objective optimization, enabling the end-to-end learning of NPSVC and its features.
By pursuing Pareto optimality,
NPSVC++ theoretically ensures feature optimality across classes, hence effectively overcoming the two issues above.
A general learning procedure via duality optimization is proposed,
based on which we provide two applicable instances, K-NPSVC++ and D-NPSVC++.
The experiments show their superiority over the existing methods and verify the efficacy of NPSVC++.

\end{abstract}

\input{intro}
\input{prel}
\input{method-motivatin}
\input{method-kernelMachine}

\input{method-dnn}

\input{exp}
\input{concl}

\section*{Impact Statement}
This paper provides a new learning scheme for nonparallel support vector classifiers, which makes these classifiers more applicable.
The applications for classification based on deep learning and kernel method would benefit from the performance enhancement of the proposed method.


\bibliography{ref}
\bibliographystyle{icml2024}

\newpage
\appendix
\onecolumn
\input{appendix}


\end{document}

%% file: notations-bold-bm.tex
\usepackage{amsthm,amssymb}
\usepackage{bm}

\newcommand{\Bc}{{\mathcal B}}

\newcommand{\Hc}{{\mathcal H}}

\newcommand{\Kc}{{\mathcal K}}

\newcommand{\Nc}{{\mathcal N}}

\newcommand{\Sc}{{\mathcal S}}

\newcommand{\Yc}{{\mathcal Y}}

\newcommand{\abf}{{\bm a}}
\newcommand{\bbf}{{\bm b}}

\newcommand{\p}{{\bm p}}

\newcommand{\ubf}{{\bm u}}
\newcommand{\vbf}{{\bm v}}
\newcommand{\w}{{\bm w}}
\newcommand{\x}{{\bm x}}
\newcommand{\y}{{\bm y}}
\newcommand{\z}{{\bm z}}

\newcommand{\A}{{\bm A}}
\newcommand{\B}{{\bm B}}

\newcommand{\D}{{\bm D}}
\newcommand{\E}{{\bm E}}

\newcommand{\G}{{\bm G}}
\newcommand{\Hbf}{{\bm H}}
\newcommand{\Ibf}{{\bm I}}

\newcommand{\K}{{\bm K}}
\newcommand{\Lbf}{{\bm L}}
\newcommand{\M}{{\bm M}}

\newcommand{\Pbf}{{\bm P}}
\newcommand{\Q}{{\bm Q}}

\newcommand{\Tbf}{{\bm T}}
\newcommand{\U}{{\bm U}}
\newcommand{\V}{{\bm V}}
\newcommand{\W}{{\bm W}}
\newcommand{\X}{{\bm X}}

\newcommand{\Z}{{\bm Z}}

\newcommand{\Rb}{{\mathbb R}}

\newcommand{\Zb}{{\mathbb Z}}

\newcommand{\blambda}{{\bm\lambda}}
\newcommand{\bbeta}{{\bm\beta}}

\newcommand{\btheta}{{\bm\theta}}
\newcommand{\bxi}{{\bm\xi}}
\newcommand{\bphi}{{\bm\phi}}
\newcommand{\bpsi}{{\bm\psi}}
\newcommand{\bpi}{{\bm\pi}}
\newcommand{\bnu}{{\bm\nu}}
\newcommand{\btau}{{\bm\tau}}

\newcommand{\bPhi}{{\bm\Phi}}
\newcommand{\bPsi}{{\bm\Psi}}

\newcommand{\T}{\top}               
\newcommand{\sto}{\mathrm{s.t.}~}

\newcommand{\sign}{\mathrm{sign}}
\newcommand{\diag}{\mathrm{diag}}

\newcommand{\onef}{{\bm 1}}

\newcommand{\tr}{Tr}

\newcommand{\wh}{\widehat}

\newcommand{\St}{\mathsf{St}}

%% file: intro.tex
\section{Introduction}
As one of the most popular classification algorithms in machine learning,
Support Vector Machine (SVM) \cite{cortes1995svm} have been investigated for decades.
The elegant margin theory behind SVM assures its powerful generalization ability \cite{mohri2018FoML}.
Due to its remarkable performance, SVM is used for various applications,
such as text classification \cite{goudjil2018novel}, and image classification \cite{chandra2021survey}.
Numerous variants emerged in response to the advent of SVM,
becoming a new development trend.
For example, Least-square SVM \cite{suykens1999LSSVM} is a representative variant of SVM still studied today \cite{Russel2023AppLSSVM}.
The nonparallel classifiers, as a collection of improved SVM, have been gaining the interest of researchers.
The basic goal of these methods is to seek a series of hyperplanes in the data space,
such that each hyperplane is close to the corresponding class and far from the others.
Each class is associated with the hyperplane closest to it.
As such, the new data will be categorized into the class of the respective nearest hyperplane.
One of the early works, \cite{Mangasarian2006GEPSVM} realized such a concept by tackling an eigensystem,
which is yet computationally expensive.
After that, \cite{Jayadeva2007twsvm} proposed Twin SVM (TWSVM), which establishes a scheme that minimizes similarity and dissimilarity loss functions to learn the classification hyperplanes.
The subsequent nonparallel classifiers, such as \cite{ArunKumar2009LSTWSVM,Shao2014NPSVM,TANVEER2019pinTWSVM},
mostly inherit this learning scheme of TWSVM.
The distinctions among these methods pertain to their loss functions and optimization algorithms.
We name these classifiers \underline{N}on\underline{P}arallel \underline{S}upport \underline{V}ector \underline{C}lassifiers (NPSVC) in this paper.

Representation learning is also one of the most important techniques in machine learning.
Typically, data is depicted as vectors in an Euclidean space, referred to as representations, or features.
The quality of the features has a substantial impact on the downstream tasks.
A basic consensus is that the discriminative features will improve the performance of classifiers \cite{xie2020mmpp}.
However, it is challenging to discover the underlying features of the intricate data.
A widely approved solution is deep learning,
which leverages the powerful expressiveness of deep neural networks (DNNs) to achieve end-to-end feature learning and classification.
DNNs with diverse architectures have demonstrated outstanding results for classification,
such as convolution neural network (CNN) \cite{Simonyan2015VGG,He2016ResNet,liu2022convnet},
and the recently popular Transformer \cite{Vaswani2017Transformer,dosovitskiy2020ViT}.
The classifier (or classification layer) is the imperative part of these DNNs that determines the learning criterion. 
Softmax regression is a typical choice for classification in deep learning, which is associated with cross-entropy loss \cite{Mao2023CE}.
SVM also usually plays the role of downstream classifier in deep learning,
and the existing DNNs with SVM have shown promising classification performance \cite{wang2019svm,diaz2020deepSVM}.
%
Although NPSVC has been extensively studied for a decade, there are still few works that investigate the deep extension of NPSVC.
Unlike most classifiers,
learning NPSVC involves multiple minimization problems,
posing the potential issues of \textit{feature suboptimality} and \textit{class dependency}.
This leaves obstacles to the development of deep extensions for NPSVCs.
Despite some attempts for deep NPSVCs, the above two issues are still not addressed.
Specifically, the model by \cite{Xie2023DeepMultiview} is not end-to-end, and thus the induced features are still suboptimal for classification;
\cite{Li2022DeepTWSVM} ignores the interaction between classes and fails to achieve the trade-off among multiple objectives.
To implement the end-to-end learning of NPSVC and guarantee its optimal representations,
it is necessary to explore a new learning strategy to address the above two issues.

\textbf{Contributions.} We develop a novel end-to-end learning framework, called \emph{NPSVC++}, which enables the incorporation of NPSVCs with representation learning through multi-objective optimization.
NPSVC++, for the first time, re-forges the learning paradigm of NPSVCs and breaks their limitations regarding feature suboptimality and class dependency,
making these classifiers more flexible and applicable.
We adopt an iterative duality optimization method to achieve Pareto optimality.
As such, in contrast to the previous methods, NPSVC++ theoretically ensures the optimality of the learned features across different objectives.
Two realizations of NPSVC++ are proposed:
K-NPSVC++ with kernel setting, and D-NPSVC++ for deep learning.
The experiments validate the effectiveness of the proposed methods.

\textbf{Notations.} We use bolded notations like $\abf$ (or $\A$) to represent vectors (or matrices).
For a matrix $\A$, its transpose is denoted as $\A^\T$.
Specially, $\A^{-\T}$ is the transpose of $\A^{-1}$ if $\A$ is invertible.
$\tr(\A)$ is the trace of matrix $\A$, i.e., the sum of diagonal elements of $\A$.
$\onef$ and $\bm 0$ are the all-ones and all-zeros vectors in proper dimension, respectively.
Denote shorthands $[n]=\{1,2,\cdots,n\}$ for any $n\in\Zb^+$ and $[\cdot]_+\triangleq\max(0,\cdot)$.
For a classification task, let $\X=[\x_1,\cdots,\x_n]\in\Rb^{m\times n}$ be the training samples,
where $\x_i\in\Rb^m$ is the $i$-th sample, and $y_i\in\Yc$ denote its label.
We denote the hypothesis function using $f(\cdot):\Rb^m\mapsto\Rb$.
Let $\Rb^{n}_+\subset\Rb^{n}$, where $\x\in\Rb^{n}_+$ means $x_i\geq0,~\forall i\in[n]$.
For two vectors $\abf,\bbf\in\Rb^n$, we say
$\abf\geq\bbf$ if $\abf-\bbf\in\Rb_+^m$,
and oppositely, $\abf\leq\bbf$ if $\bbf-\abf\in\Rb_+^m$.
Define the Euclidean projection of $\x$ to a set $\Sc$ as
$\bpi_{\Sc}(\x)=\min_{\y\in\Sc}\|\x-\y\|^2$.
    


%% file: prel.tex
\section{Brief Review of NPSVC}\label{sec:prel}
NPSVCs indicate a series of improved models of SVM.
Nevertheless, their learning scheme differs from SVM and many mainstream classifiers like softmax regression.
NPSVCs classify the data with multiple non-parallel hyperplanes.
For binary-class data, $\Yc=\{-1,1\}$,
an NPSVC seeks a pair of hyperplanes, namely the positive hyperplane $f_+(\x)=\w_+^\T\x+b_+=0$ and negative hyperplane $f_-(\x)=\w_-^\T\x+b_-=0$.
The data will be classified into the class corresponding to the nearest hyperplane.
Thus, binary-class NPSVC applies the following prediction rule.
\begin{equation*}
    h(\x')=\sign\left(\frac{|f_-(\x')|}{\|\w_-\|}-\frac{|f_+(\x')|}{\|\w_+\|}\right).
\end{equation*}
This means the positive hyperplane should be close to the positive samples ($y_i=1$) and far away from negative samples ($y_i=-1$),
and the negative hyperplane behaves oppositely.
As an intuitive explanation,
each hyperplane draws a profile of each class,
and each class is attached to the closest hyperplane.
To this end, NPSVC consists of two problems that optimize different hyperplanes:
\begin{equation}
    \begin{aligned}
    &\min_{f_+}R(f_+)+\sum_{y_i>0}\ell_s(f_+(\x_i))+c\sum_{y_i<0}\ell_d(f_+(\x_i)),\\
    &\min_{f_-}R(f_-)+\sum_{y_i<0}\ell_s(f_-(\x_i))+c\sum_{y_i>0}\ell_d(f_-(\x_i)),
    \end{aligned}\label{eq:NPSVC-bin}
\end{equation}
where $c>0$ is hyperparameter, $R(\cdot)$ is a regularizer,
and $\ell_s$ and $\ell_d$ denote similarity and dissimilarity loss, respectively.
Twin support vector machine (TWSVM) \cite{Jayadeva2007twsvm} is one of the most popular models of NPSVC,
which utilizes $\ell_s(a)=a^2/2$ and $\ell_d(a)=[1-a]_+$.
Table \ref{tab:npsvc} lists some typical NPSVC methods and their loss functions.

Now we consider a multi-class setting with $\Yc=[K]$.
One-versus-rest (OVR) strategy can be applied to implement multi-class NPSVC \cite{Xie2013OVRTWSVM}.
It seeks $K$ hyperplanes such that the samples from $l$-th class are close to $l$-th hyperplane $f_l(\x)=0$, where $f_l(\x)=\w_l^\T\x+b_l$, and other samples are far from it.
Analogous to the binary-class case, the prediction rule is to match the nearest hyperplane: 
\begin{equation*}
    h(\x')=\arg\min_{l\in[K]}\frac{|f_l(\x')|}{\|\w_l\|}.
\end{equation*}
Accordingly, it formulates the training criterion as:
\begin{equation}
    \min_{f_l}R(f_l)+\sum_{y_i=l}\ell_s(f_l(\x_i))+c\sum_{y_i\not=l}\ell_d(f_l(\x_i)),~ l\in[K].\label{eq:NPSVC}
\end{equation}
It should be noted that \eqref{eq:NPSVC} consists of $K$ problems.
Unless explicitly stated otherwise, the remaining analysis in this work uses the OVR strategy for the multi-class setting.
\begin{table}[t]
    \renewcommand\arraystretch{1.05}
    \centering
    \caption{Representative methods of NPSVC. Here, $s$ denotes the margin size, generally $s=1$.
    $\epsilon, \tau,\gamma$ and $t$ are the hyperparameters.
    Generally, minimizing $\ell_s(f_l(\x_i))$ makes $f_l(\x_i)\to 0$,
    while minimizing $\ell_d(f_l(\x_i))$ imposes $f_l(\x_i)\geq s$ or $f_l(\x_i)\approx s$.}
    \scalebox{0.8}{
    \begin{tabular}{l|l|l}
        \hline\hline
        References of method & $\ell_s(a)$    & $\ell_d(a)$ \\
        \hline
        \cite{Jayadeva2007twsvm} & $a^2/2$ & $[s-a]_+$\\
        \hline
        \cite{ArunKumar2009LSTWSVM} & $a^2/2$ & $(s-a)^2/2$ \\
        \hline
        \cite{Tian2014NPSVM} & $[|a|-\epsilon]_+$ & $[s-a]_+$\\
        \hline
        \cite{TANVEER2019pinTWSVM} & $a^2/2$ &$\begin{cases}
            s-a & a\leq s\\
            -\tau(s-a) & a > s
        \end{cases}$
        \\
        \hline
        \cite{Yuan2021CTWSVM} & $1-e^{-\gamma|a|}$ & $[s-a]_+$\\
        \hline
        \cite{Zhang2023RMTBSVM} & $\min(|a|,\epsilon)$ & $[s-a]_+$\\
        \hline\hline
    \end{tabular}
    }
    \label{tab:npsvc}
    \vspace{-15pt}
\end{table}

%% file: method-motivatin.tex
\begin{figure*}
    \centering
    \includegraphics[height=0.22\textheight]{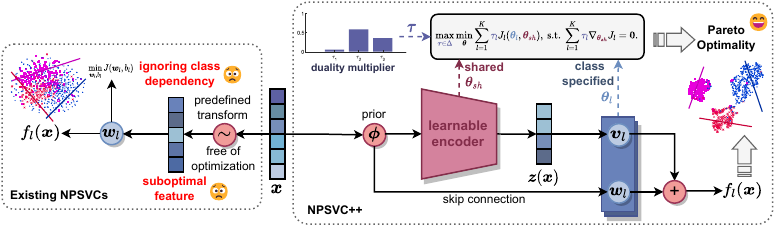}
    \caption{\textbf{Left:} NPSVC usually uses learning-freed feature transformation and optimizes the hyperplane of each class independently, resulting in feature suboptimality and disregarding class dependency.
    \textbf{Right:} NPSVC++ stems from multi-objective optimization, whose goal is to achieve the Pareto optimality.
    Thus, feature optimality across different classes is ensured, tackling the issues of NPSVCs.}
    \label{fig:overview}
    \vspace{-12pt}
\end{figure*}

\section{Methodology of NPSVC++}
\subsection{Motivation: Issues in NPSVCs}
The current non-parallel classifiers mostly follow the framework \eqref{eq:NPSVC},
where $f_l$ is independent of each other.
They exploit various loss functions to develop the model's capacity and robustness.
However, the existing NPSVCs still encounter two common issues:

\textbf{Feature suboptimality.}
NPSVC assumes the features of each class can be illustrated by the associated hyperplanes.
Thus the quality of the input features will affect the performance of NPSVC.
Some existing works have attempted to enhance NPSVC via feature processing.
For example, the feature selection methods are combined with NPSVCs \cite{Gao2021L1TWSVM,Bai2014FeaSelTWSVM}.
However, they usually focus on eliminating the unimportant features in raw data rather than improving the discrimination.
On the other hand, \cite{Xie2023DeepMultiview} extracts the feature with a trained autoencoder network.
Though informative features are obtained,
these features lack supervision information and are thus unsuitable for NPSVC.
In fact, it optimizes features and classifiers in distinctive criteria, resulting in feature suboptimality. 
Thus, it is expected to design an effective end-to-end model that integrates NPSVC with representation learning.

\textbf{Class dependency.}
The second issue in NPSVC is how to make the learning be ``class-collaborative''.
We have mentioned that the $K$ problems in NPSVC \eqref{eq:NPSVC} are optimized independently.
In other words,
the learning of each hyperplane is agnostic to others.
Hence, the correlation between different classes is ignored, which likely results in inadequate performance.
For example,
consider an image classification task with three classes:
``rabbit'', ``tiger'' and ``leopard''.
Intuitively, tigers and leopards likely have some similar characteristics as they both belong to the Felidae family, while NPSVCs likely fail to recognize this distinction across different classes due to their learning mechanism.

To address these two issues,
\cite{Li2022DeepTWSVM} has shown a valuable attempt that integrates representation learning into NPSVC.
It constructs a deep neural network based on TWSVM with all classes sharing the same encoder (i.e., feature extractor).
Thereby, the inherent dependency between different classes can be captured by a single encoder,
and thus more informative features will be obtained.
But a new challenge arises.
As the shared encoder is introduced, the objective functions in \eqref{eq:NPSVC} become dependent.
The previous training way in NPSVC, which treats different objectives independently, becomes improper.
Therefore, a new strategy to handle the class dependency is demanded.
However, \cite{Li2022DeepTWSVM} only trivially minimizes the sum of all objectives,
which disregards their significance and results in deficiency.
The underlying reason is that it fails to reach the trade-off among different objectives and the optimality of each objective is uncertain,
while no work has investigated such optimality in NPSVC with dependent objectives yet.

Accordingly, two essential problems emerge:
``\emph{how to devise a representation learning criterion for NPSVC}?'', and ``\emph{how to optimize multiple dependent objectives derived from this criterion}?''
To answer these,
we propose a framework composing representation learning and nonparallel classifiers based on multi-objective optimization, namely NPSVC++.

\subsection{NPSVC++: From Multi-objective Optimization}
Recall the optimization model of NPSVC \eqref{eq:NPSVC}.
Denote $J_l^\text{NP}$ as the $l$-th objective function of \eqref{eq:NPSVC}.
Then NPSVC can be reformulated as a multi-objective optimization problem:
\begin{equation}
    \min_{\{\w_l,b_l\}_{l=1}^{K}}\bm J\triangleq[J^{\text{NP}}_1(\w_1,b_1),\cdots,J^{\text{NP}}_K(\w_K,b_K)],\label{eq:NPSVC-mul-obj}
\end{equation}
which means that all objective functions $J_l^{\text{NP}}$ are minimized simultaneously.
Indeed, in NPSVC, each objective $J_l^{\text{NP}}$ has no shared variables, meaning that the objectives are mutually independent.
Thus, problem \eqref{eq:NPSVC-mul-obj} is equivalent to \eqref{eq:NPSVC}, which means each objective can be minimized independently.
Alternatively, we consider a more general multi-objective optimization framework with shared variables:
\begin{equation}
    \min_{\btheta}\bm J(\btheta)\triangleq\left[J_1(\btheta_1,\btheta_{sh}),\cdots,J_K(\btheta_K,\btheta_{sh})\right],\label{eq:NPSVCPlus-vecopt}
\end{equation}
where $\btheta_l$ denotes the variables specific to the $l$-th class and $\btheta_{sh}$ represents the shared variables for all classes.
Let $\btheta=\{(\btheta_l)_{l=1}^K,\btheta_{sh}\}$ denote all optimization variables.

We now formulate NPSVC++ following the framework \eqref{eq:NPSVCPlus-vecopt}.
First, similar to NPSVC,
the objective function of each class $J_l$ is formulated as:
\begin{equation}
    J_l\triangleq\sum_{y_i=l}\ell_s(f_l(\x_i))+c\sum_{y_i\not=l}\ell_d(f_l(\x_i))+R(f_l),\label{eq:ahtwsvm-Jl}\\
\end{equation}
where, distinctive from NPSVC, the hypothesis function of NPSVC++ is defined as\footnote{We omit the bias term $b_l$ for convenient discussion. One can append an additional $1$ after features to implicitly include the bias.}:
\begin{equation}
    f_l(\x)=\langle\w_l,\bphi(\x)\rangle+\vbf_l^\T\z(\x)\label{eq:npsvcpp-f}.
\end{equation}
Here, $\w_l,\vbf_l$ are class-specific weights,
$\bphi(\cdot)$ is a prior transformation (which is predefined),
and $\z(\cdot)$ is the encoder shared by all classes.
Hypothesis \eqref{eq:npsvcpp-f} exhibits a skip connection structure,
which aims to preserve the raw information $\bphi(\x)$ and avoid classfier's degradation.
Recalling \eqref{eq:NPSVCPlus-vecopt}, in NPSVC++, the class-specific variable is
$\btheta_l=(\w_l,\vbf_l)$ and the shared variable $\btheta_{sh}$ is the parameters of encoder $\z(\cdot)$.
The major challenge in NPSVC++ stems from the shared variable $\btheta_{sh}$.
Since the change of the shared variable affects all objective functions,
its optimization would entail intricate interactions or potential conflicts among various objectives \cite{Multitask2018Sener}.
Hence, a fundamental goal is to achieve a compromise among different objectives.
To this end, NPSVC++ endeavors to pursue the \textit{Pareto optimal solution}.
We first introduce the definition for explanation.
\begin{definition}[Pareto optimality \cite{boyd2004CVXOPT}]
    A solution $\btheta^\star$ to \eqref{eq:NPSVCPlus-vecopt} is Pareto optimal,
    if there exists no $\btheta$ such that $J_l(\btheta)\leq J_l(\btheta^\star)$ for all $l\in[K]$ and $\bm J(\btheta)\not=\bm J(\btheta^\star)$.
\end{definition}
The Pareto optimality means there exists no better solution that can further improve all objectives.
In NPSVC++, Pareto optimality is the key to resolving the issues of feature suboptimality and class dependency,
since it implies ``the features and classifiers are \textit{optimal} for \textit{all classes}'', while the previous methods have no such theoretical guarantee.
Generally, Pareto optimality is achieved by the Pareto stationarity.
\begin{definition}[Pareto stationarity \cite{MGDA2012}]\label{def:ParetoStation}
    A solution to \eqref{eq:NPSVCPlus-vecopt} is Pareto stationary if existing $\btau\in\Rb_+^K$ satisfies:
    \begin{equation}
        \sum_{l=1}^K\tau_l\nabla_{\btheta_{sh}}J_l(\btheta_l,\btheta_{sh})=\bm0,~\sum_{l=1}^K\tau_l=1,~\tau_l\geq 0.
    \end{equation}
\end{definition}
Pareto stationarity is a necessary condition of the Pareto optimality \cite{MGDA2012}.
Therefore, it is essential to seek an appropriate Pareto stationary point and the corresponding $\btau$.
Inspired by \cite{Giagkiozis2015MultiObj,momma2022multiobj},
we consider the following weighted Chebyshev decomposition problem with $L_\infty$-metric:
\begin{equation}
    \min_{\rho,\btheta}~\rho,\quad\sto \varpi_lJ_l(\btheta_l,\btheta_{sh})\leq \rho,\forall l\in[K],\label{eq:Wcheb-prob}
\end{equation}
where $\varpi_l>0$ denotes the predefined preference for $l$-th objective.
For simplification, we let $\varpi_l=1,\forall l\in[K]$ in this paper.
Problem \eqref{eq:Wcheb-prob} minimizes all objective functions via suppressing their upper bound $\rho$.
To solve it, we first write its dual problem, which is formulated as a linear programming problem imposing the Pareto stationarity constraint.
This result is demonstrated by the following theorem.
\begin{theorem}\label{thm:mul-obj}
    The dual problem of \eqref{eq:Wcheb-prob} is
    \begin{equation}
        \begin{aligned}
            \max_{\btau}\min_{\btheta}&\sum_{l=1}^K\tau_lJ_l(\btheta_l,\btheta_{sh}),\\
            \sto&\sum_{l=1}^K\tau_l\nabla_{\btheta_{sh}}J_l=\bm0,~
            \onef^\T\btau=1,~\tau_l\geq 0,
        \end{aligned}
        \label{eq:dual-Wcheb-prob}
    \end{equation}
    where $\btau\in\Rb_+^K$ denotes Lagrange multipliers.
\end{theorem}
We provide the proof in Appendix \ref{app:proof-thm-mul-obj}.
The key to training NPSVC++ is to solve the max-min dual problem \eqref{eq:dual-Wcheb-prob},
leading to the following iterative optimization procedure:
\begin{tcolorbox}[left=0mm,right=1.2mm,top=0mm,bottom=2.5mm]
\begin{align}
    \btheta_{t+1}&=\arg\min_{\btheta} \btau_t^\T\bm J(\btheta),\label{eq:theta-step}\\
    \btau_{t+1}&=\arg\max_{\btau\in\Delta} \btau^\T\bm J(\btheta_{t+1}),~\sto\bm\nabla_{sh}\btau=\bm0,\label{eq:tau-step}
\end{align}
\end{tcolorbox}
where matrix $\bm\nabla_{sh}=[\nabla_{\btheta_{sh}}J_1,\cdots,\nabla_{\btheta_{sh}}J_K]$,
and $\Delta=\{\btau\in\Rb^{K}_+:\onef^\T\btau=1\}$ denotes the probability simplex.
The fundamental structure of NPSVC++ is illustrated in Figure \ref{fig:overview}. 
In the following sections,
we will elaborate on the learning of NPSVC++ with two typical settings,
i.e., kernel machine and deep neural networks.

%% file: method-kernelMachine.tex
\section{Instantiation}
\subsection{NPSVC++ as Kernel Machine: K-NPSVC++}
Let prior $\bphi(\cdot)$ be the induced mapping of kernel $\Kc(\cdot,\cdot)$, which is assumed positive-definite,
and $\Hc$ is the corresponding reproducing kernel Hilbert space (RKHS).
Denote $\langle\cdot,\cdot\rangle$ the inner product operator in $\Hc$.
We define Stiefel manifold $\St(\Hc,d)=\{\Pbf\in\Hc^d:\Pbf^\T\Pbf=\Ibf\}$.
In some unambiguous contexts,
we directly write notation $\St$ for simplification.

In the following discussion,
we denote $\bphi(\X)=\left[\bphi(\x_i)\right]_{i=1}^n$ the feature matrix,
and $\K\in\Rb^{n\times n}$ the kernel matrix with each entry $K_{ij}=\Kc(\x_i,\x_j)$.
We parameterize $\z(\x)=\Pbf^\T\bphi(\x)$ with a projection $\Pbf\in\St(\Hc,d)$.
Therefore,
\begin{equation}
\begin{aligned}
    f_l(\x)&=\langle\w_l,\bphi(\x)\rangle+\vbf_l^\T\z(\x)\\
    &=\langle\w_l+\Pbf\vbf_l,\bphi(\x)\rangle=\langle\ubf_l,\bphi(\x)\rangle,
    \label{eq:fx-Pform}
\end{aligned}
\end{equation}
where we let $\ubf_l=\w_l+\Pbf\vbf_l$.
This substitution will simplify the subsequent optimization,
and the optimization variables thus become $\btheta=\{\U,\V,\Pbf\}$,
where $\U=[\ubf_1,\cdots,\ubf_K]\in\Hc^K$ and $\V=[\vbf_1,\cdots,\vbf_K]\in\Rb^{d\times K}$.
In this instance,
we inherit the loss of TWSVM,
i.e., $\ell_s(a)=a^2/2$ and $\ell_d(a)=[1-a]_+$,
and use the following regularizer:
\begin{equation*}
\begin{aligned}
    R(\w_l,\vbf_l,\z)&=\frac{r_1}{2}\|\w_l\|^2+\frac{r_2}{2}\|\vbf_l\|^2\\
    &+\frac{\mu}4\sum_{i,j}\|\z(\x_i)-\z(\x_j)\|^2G_{ij}, 
\end{aligned}
\end{equation*}
where $r_1,r_2$ and $\mu$ are positive hyperparameters.
The third term, known as the manifold regularization, aims to capture the locality structure across the data,
where the graph adjacency matrix $\G\in\Rb^{n\times n}$ describes the pairwise relation of the data.
Denote $\D$ a diagonal matrix with the $i$-th diagonal element $D_{ii}=\sum_{j=1}^nG_{ij}$
and $\Lbf=\D-\A$ the Laplacian matrix.
Then the $l$-th objective function is reformulated as
\begin{equation*}
    \begin{aligned}
    J_l&=\frac12\sum_{y_i=l}\langle\ubf_l,\bphi(\x_i)\rangle^2+c\sum_{y_i\not=l}[1-\langle\ubf_l,\bphi(\x_i)\rangle]_+\\
       &+\frac{r_1}2\|\ubf_l-\Pbf\vbf_l\|^2+\frac{r_2}2\|\vbf_l\|^2+\frac{\mu}{2}\tr(\Z\Lbf\Z^\T),\notag\\
    \end{aligned}
\end{equation*}
where $\Z=\Pbf^\T\bphi(\X)$ and the regularization of $\w$ is written as $\|\w_l\|^2=\|\ubf_l-\Pbf\vbf_l\|^2$.
By Representer Theorem \cite{scholkopf2002kernel},
there exists $\A=[\alpha_{ij}]\in\Rb^{n\times d}$ where $\A^\T\K\A=\Ibf$ and $\B=[\beta_{il}]\in\Rb^{n\times K}$,
such that 
\begin{equation}
    \ubf_l^*=\sum_{i=1}^n\beta_{il}\bphi(\x_i),\quad \p_j^*=\sum_{i=1}^n\alpha_{ij}\bphi(\x_i).
\end{equation}
Then the objective function is reformulated as
\begin{align*}
    J_l&=\frac12\|\K_{l}^\T\bbeta_l\|^2+c\sum_{y_i\not=l}\left[1-\bbeta_l^\T\bm\kappa_i\right]_++\frac{r_2}2\|\vbf_l\|^2\notag\\
        &+\frac{r_1}{2}\|\bphi(\X)\bbeta_l-\bphi(\X)\A\vbf_l\|^2+\frac\mu2\tr(\A^\T\K\Lbf\K\A),
\end{align*}
where $\bm\kappa_i$ is the $i$-th column of $\K$ and $\K_{l}=[\bm\kappa_i]_{y_i=l}$.
We employ a further substitution to simplify the computation.
Instead of directly optimizing $\A$ and $\B$,
we optimize $\wh\Pbf=\bPsi^\T\A$ and $\wh\U=\bPsi^\T\B$, 
where $\bPsi$ suffices the decomposition $\K=\bPsi\bPsi^\T$.
Matrix $\bPsi^\T$ can be seen as the empirical version of $\bphi(\X)$,
and thus $\wh\Pbf\in\Rb^{n\times d}$ and $\wh\U\in\Rb^{n\times K}$ serve as the empirical version of $\Pbf\in\Hc^d$ and $\U\in\Hc^K$, respectively.
We generally set $\bPsi$ as the Cholesky factor of $\K$, which ensures the invertibility of $\bPsi$.
In this way, the substitution is bijective and $\A=\bPsi^{-\T}\wh\Pbf$ and $\B=\bPsi^{-\T}\wh\U$.
The training will benefit from this substitution strategy in two aspects:
\vspace{-8pt}
\begin{itemize}
    \item It avoids handling the complex constraint $\A^\T\K\A=\Ibf$.
    Instead, the constraint becomes $\wh\Pbf^\T\wh\Pbf=\Ibf$, or equivalently, $\wh\Pbf\in\St(\Rb^{n},d)$,
    allowing us to employ existing efficient Stiefel optimization methods.
    \vspace{-4pt}
    \item The formulation $\K=\bPsi\bPsi^\T$ is consistent with the linear kernel $\Kc(\x_i,\x_j)=\x_i^\T\x_j$.
    Hence,
    one can directly set $\bPsi=\X^\T$ to implement linear classification without computing the coefficients $\A,\B$.
\end{itemize}
\vspace{-8pt}
Now, with a series of transformations,
the eventual objective function becomes:
\begin{align*}
    J_l&=\frac12\|\bPsi_{l}\wh\ubf_l\|^2+c\sum_{y_i\not=l}\left[1-\wh\ubf_l^\T\bm\bpsi_i\right]_++\frac{r_2}2\|\vbf_l\|^2\\
        &+\frac{r_1}{2}\|\wh\ubf_l-\wh\Pbf\vbf_l\|^2+\frac\mu2\tr(\wh\Pbf^\T\bPsi^\T\Lbf\bPsi\wh\Pbf),
\end{align*}
where $\bpsi_i\in\Rb^n$ is the $i$-th row of $\bPsi$.
We apply the alternative minimization strategy to implement minimization step \eqref{eq:theta-step},
which is elaborated as follows:

$\U$ \textbf{step}.
In this step, we fix $\wh\V$ and $\wh\Pbf$ to optimize $\wh\ubf_l$ for each $l\in[K]$.
We first compute three auxiliary matrices:
\begin{equation}
\begin{aligned}
    \Q^{(l)}_{\Psi,\Psi}&=\frac{1}{r_1}\left[\Ibf-\bPsi_{l}^\T(r_1\Ibf+\bPsi_{l}\bPsi_{l}^\T)^{-1}\bPsi_{l}\right],\\
    \Q^{(l)}_{K,\Psi}&=\bPsi_{-l}^\T\Q^{(l)}_{\Psi,\Psi},\quad \Q^{(l)}_{K,K}=\bPsi_{-l}^\T\Q^{(l)}_{\Psi,\Psi}\bPsi_{-l},
\end{aligned}
\label{eq:kNPSVCPlus-Qmat-compute}
\end{equation}
where $\bPsi_l=[\bpsi_i]_{y_i=l}$ denotes those rows in $\bPsi$ of $l$-th class,
and $\bPsi_{-l}$ corresponds to the other rows.
Then we have the following closed-form solution:
\begin{equation}
    \wh\ubf_l=r_1\Q^{(l)}_{\Psi,\Psi}\wh\Pbf\vbf_l+\Q^{(l)}_{K,\Psi}\blambda_l,\label{eq:kNPSVCPlus-U}
\end{equation}
where $\blambda_l$ denotes the Lagrange multiplier and is obtained by the following dual problem:
\begin{equation}
    \min_{\bm0\leq\blambda_l\leq c\onef}\frac12\blambda_l^\T\Q^{(l)}_{K,K}\blambda_l+(\Q^{(l)}_{\Psi,K}\wh\Pbf\vbf_l-\onef)^\T\blambda_l.\label{eq:kNPSVCPlus-U-qp}
\end{equation}
We leave the detailed deduction in Appendix \ref{app:u}.

$\V$ \textbf{step}.
Fixing $\wh\U$ and $\wh\Pbf$, we have
\begin{align*}
    &\min_{\vbf_l}\frac{r_1}{2}\|\wh\ubf_l-\wh\Pbf\vbf_l\|^2+\frac{r_2}{2}\|\vbf_l\|^2.
\end{align*}
Concisely, setting the derivative of $\vbf_l$ to zero, we obtain
\begin{equation}
    \vbf_l=\frac{r_1}{r_1+r_2}\wh\Pbf^\T\ubf_l.
\end{equation}

$\Pbf$ \textbf{step}.
We optimize $\Pbf$ by the following problem:
\begin{equation*}
    \min_{\wh\Pbf^\T\wh\Pbf=\Ibf}{r_1}\sum_{l=1}^K\tau_l\|\wh\ubf_l-\wh\Pbf\vbf_l\|^2+{\mu}\tr(\wh\Pbf^\T\bPsi\Lbf\bPsi^\T\wh\Pbf).\label{eq:ahtwsvm-probP}
\end{equation*}
With some simple algebra, the problem becomes
\begin{equation}
    \min_{\wh\Pbf^\T\wh\Pbf=\Ibf}\mu\tr(\wh\Pbf^\T\bPsi^\T\Lbf\bPsi\wh\Pbf)-2r_1\tr(\wh\Pbf^\T\wh\U\Tbf\V^\T),\label{eq:solve-P}
\end{equation}
where $\Tbf=\diag(\btau)$.
Problem \eqref{eq:solve-P} is a quadratic programming problem over Stiefel manifold $\St(\Rb^n,d)$.
Here we leverage an efficient solver, i.e., generalized power iteration (GPI) \cite{nie2017GPI}.
Note that \eqref{eq:solve-P} is equivalent to
\begin{equation}
    \max_{\Pbf^\T\Pbf=\Ibf}\tr(\Pbf^\T\Hbf\Pbf)+2\tr(\Pbf^\T\E),\label{eq:gpi-max}
\end{equation}
where $\E=r_1\U\Tbf\V^\T$ and $\Hbf=\sigma\Ibf-\mu\bPsi^\T\Lbf\bPsi$ with a predefined $\sigma>0$
such that $\Hbf$ is positive definite.
The large $\sigma$ might cause slow convergence \cite{nie2017GPI}.
Thus, we can set $\sigma=1+\mu\sigma_{\max}(\bPsi^\T\Lbf\bPsi)$,
where $\sigma_{\max}(\cdot)$ gives the largest singular value of the input matrix.
GPI generate a series of $\Pbf_{t}$ converging to the minimum of \eqref{eq:gpi-max} via iterative process
$\Pbf_{t+1}=\bpi_\St(\Hbf\Pbf_{t}+\E)$. 
Specifically, the Euclidean projection of $\A$ onto Stiefel manifold is $\bpi_\St(\A)=\bm\Pi\bm\Gamma^\T$
with the compact SVD: $\bm\Pi\bm\Sigma\bm\Gamma^\T=\A$ \cite{zou2006spca}.

$\btau$ \textbf{step}.
In this step, we solve the maximization problem \eqref{eq:tau-step} to update $\btau$.
We first consider the equality constraint $\bm\nabla_{sh}\btau=\bm 0$ in \eqref{eq:tau-step} with the shared variable $\btheta_{sh}=\wh\Pbf$.
Note that $\Pbf$ is constrained in the Stiefel manifold $\St(\Rb^n,d)$.
Therefore, we should utilize the corresponding Riemannian gradient instead of the gradient in Euclidean space.
According to \cite{wen2013Stiefel}, the Riemannian gradient over the Stiefel manifold is
\begin{equation*}
    \nabla^{\St}_{\wh\Pbf} J_l=(\nabla_{\wh\Pbf} J_l\wh\Pbf^\T-\wh\Pbf\nabla_{\wh\Pbf} J_l^\T)\wh\Pbf,
\end{equation*}
where $\nabla_{\wh\Pbf} J_l=\mu\bPsi^\T\Lbf\bPsi\wh\Pbf-r_1\ubf_l\vbf_l^\T$.
On the other hand, there probably exists no $\btau$ suffices for the linear constraints $\sum_{l=1}^K\tau_l\nabla_{\wh\Pbf}^\St J_l=\bm0$ as it is an overdetermined equation.
Therefore, we optimize the relaxed problem of \eqref{eq:tau-step} instead:
\begin{equation}
    \min_{\btau\in\Delta}\frac12\left\|\sum_{l=1}^K\tau_l\nabla_{\wh\Pbf}^\St J_l\right\|^2-\gamma\sum_{l=1}^K\tau_lJ_l,\label{eq:NPSVCPlus-tau}
\end{equation}
where $\gamma>0$ is a hyperparameter.
This is a QPP with probability simplex constraint.
An efficient solver is the double-coordinate descent method \cite{beck2014DoubleCoordinate}.
Remarkably, the following theorem demonstrates a valuable property of the optimal $\btau$.
\begin{theorem}[\cite{MGDA2012,momma2022multiobj}]
    If $\gamma=0$, let $\btau^\star$ the optimal solution to \eqref{eq:NPSVCPlus-tau}, and we have:
    \begin{enumerate}
        \item if the minimum is zero, then $\wh\Pbf$ is a Pareto stationary point;
        \item otherwise, $\Delta\Pbf=-\sum_{l=1}^K\tau^\star_l\nabla_\Pbf^\St J_l$ is the direction that descends all $J_l,~\forall l\in[K]$.
    \end{enumerate}
\end{theorem}
In a nutshell,
the optimal solution to \eqref{eq:NPSVCPlus-tau} yields a direction that reduces all objectives when $\gamma=0$,
which is a by-product of updating $\btau$.
Therefore, we can utilize the projected gradient descent step to further update $\wh\Pbf$:
\begin{equation*}
    \wh\Pbf:=\bpi_{\St}\left(\wh\Pbf-\eta\sum_{l=1}^K\tau_l^{\star}\nabla_{\wh\Pbf}^\St J_l\right),
\end{equation*}
where $\eta>0$ is the learning rate.
In practice, we should set $\gamma>0$ and thus the presence of the dual maximization term in \eqref{eq:NPSVCPlus-tau} will yield a perturbation in this descending direction.
Hence, we should determine a proper $\gamma$ to balance the two terms in \eqref{eq:NPSVCPlus-tau} and thereby achieve a trade-off between Pareto stationarity and dual maximization.

The training of K-NPSVC++ is summarized as Algorithm \ref{alg:K-NPSVCPP} in Appendix \ref{app:alg}.
In the testing phase,
the following decision function is used:
\begin{equation}
    \begin{aligned}
        h(\x')&=\arg\min_{l\in[K]}\frac{|f_l(\x')|}{\sqrt{\|\w_l\|^2+\|\vbf_l\|^2}}\\
        &=\arg\min_{l\in[K]}\frac{|f_l(\x')|}{\sqrt{\|\wh\ubf_l-\wh\Pbf\vbf_l\|^2+\|\vbf_l\|^2}},
    \end{aligned}
\end{equation}
where $f_l(\x)=\sum_{i=1}^n\beta_{il}\Kc(\x_i,\x)$ and $\bbeta_l=\bPsi^{-\T}\wh\ubf_l$.
In particular, if the linear kernel is used, one can directly compute $f_l(\x)=\wh\ubf_l^\T\x$ without accessing $\bbeta_l$.

%% file: method-dnn.tex
\subsection{NPSVC++ in Deep Learning: D-NPSVC++}
To employ NPSVC++ by deep learning,
we first note that the hypothesis \eqref{eq:npsvcpp-f} coincides the structure of skip connection \cite{He2016ResNet}. Therefore, it can be rewritten as
\begin{equation}
    f_l(\x)=\begin{bmatrix}
        \w_l\\\vbf_l
    \end{bmatrix}^\T\begin{bmatrix}
        \bphi(\x)\\ \z(\x)
    \end{bmatrix}=\tilde\w_l^\T\tilde\z(\x),
\end{equation}
which can be constructed as a linear layer.
The prior encoder $\phi(\cdot)$ could be a pretrained network (e.g., ResNet34 in our experiments).
For simplification, We parameterize the encoder with a multi-layer perception (MLP): $\z(\x)=\mathsf{MLP}(\bphi(\x))$.

\newcommand\topone[1]{\colorbox[RGB]{255, 250, 166}{\textbf{#1}}}
\newcommand\topsec[1]{\colorbox[RGB]{246, 214, 159}{\textit{#1}}}
\newcommand\verysig{$_{\ddagger}$}
\newcommand\somesig{$_{\dagger}$}
\newcommand\nosig{$_{\phantom{\dagger}}$}
\begin{table*}[t]
    \renewcommand\arraystretch{1.12}
    \centering
    \caption{Classification performance of SVMs. The number of training samples, testing samples, dimensions, and classes of each dataset are shown in the second column.
    ``$\ddagger$'' and ``$\dagger$'' denote the significance level of 0.05 and 0.1, respectively. The best results are bolded.}
    \vspace{2pt}
    \scalebox{0.84}{
    \begin{tabular}{ll|ccccc|c}
    \hline\hline
        \multirow{2}*{Dataset} & \multirow{2}*{$(n_{tr},~~n_{te},~~~m,\quad ~K)$} & \multicolumn{6}{c}{Methods}\\ \cline{3-8}
        & &SVM & TWSVM & NPSVM & pinTWSVM & RMTBSVM & K-NPSVC++\\\hline
        CHG & (540, ~~360, ~~8000, 10) & 82.08$\pm$2.01\verysig  & 85.75$\pm$2.41\verysig  & 86.83$\pm$3.14\verysig  & 87.42$\pm$2.61\somesig  & 87.28$\pm$2.44\verysig  & \textbf{88.14}$\pm$2.21  \\ 
        BinAlpha & (842, ~~562, ~~320, ~~36) & 71.10$\pm$2.19\nosig  & 68.06$\pm$1.86\verysig  & 69.59$\pm$1.63\verysig  & 64.50$\pm$1.46\verysig  & 69.68$\pm$2.42\verysig  & \textbf{71.28}$\pm$1.79  \\ 
        20News & (9745, 6497, 100, ~~4) & 79.31$\pm$0.43\verysig  & 80.33$\pm$0.37\verysig  & 81.86$\pm$0.42\nosig  & 80.11$\pm$0.32\verysig  & 80.72$\pm$0.44\verysig  & \textbf{81.92}$\pm$0.36  \\ 
        Pendigits & (6595, 4397, 16, ~~~~10) & 91.75$\pm$0.26\verysig  & 99.35$\pm$0.10\nosig  & 99.38$\pm$0.09\nosig  & 95.45$\pm$0.48\verysig  & 99.04$\pm$0.16\verysig  & \textbf{99.39}$\pm$0.08  \\ 
        DNA & (1200, 800, ~~180, ~~3) & 94.34$\pm$0.67\verysig  & 95.63$\pm$0.73\nosig  & 95.55$\pm$0.70\nosig  & 95.51$\pm$0.72\nosig  & \textbf{95.68}$\pm$0.75\nosig  & 95.63$\pm$0.56  \\ 
        USPS & (4374, 2917, 256, ~~10) & 96.63$\pm$0.30\verysig & 97.51$\pm$0.32\verysig & 97.87$\pm$0.32\verysig & 98.06$\pm$0.26\verysig & 97.86$\pm$0.22\verysig & \textbf{98.08}$\pm$0.24\\\hline\hline
    \end{tabular}
    }
    \label{tab:exp-svm}
    \vspace{-10pt}
\end{table*}

The training of D-NPSVC++ still follows the criterion \eqref{eq:dual-Wcheb-prob}.
In the first step \eqref{eq:theta-step},
we optimize the network parameters with the back-propagation algorithm,
following the criterion: $\min_{\btheta}\sum_{l=1}^K\tau_lJ_l$.
Similar to K-NPSVC++, the second step \eqref{eq:tau-step} solves the following problem:
\begin{equation}
    \min_{\btau\in\Delta}\frac12\left\|\sum_{l=1}^K\tau_l\nabla_{\btheta_{sh}}J_l\right\|^2-\gamma\sum_{l=1}^K\tau_lJ_l.\label{eq:d-npsvcpp-tau}
\end{equation}
However, the shared parameter $\btheta_{sh}$ (i.e., the parameter of $\z(\cdot)$) is generally quite high-dimensional.
Therefore, it is of high possibility that $\bm0\not\in\mathrm{span}(\{\nabla_{\btheta_{sh}}J_l\}_{l=1}^K)$,
which makes it difficult to minimize the first term of \eqref{eq:d-npsvcpp-tau}.
To alleviate this problem, we first leverage the chain rule and the property of the norm and obtain:
\begin{equation}
    \left\|\sum_{l=1}^K\tau_l\nabla_{\btheta_{sh}}J_l\right\|\leq \left\|\frac{\partial \Z_\Bc}{\partial \btheta_{sh}}\right\|\left\|\sum_{l=1}^K\tau_l\nabla_{\Z_\Bc}J_l\right\|.
\end{equation}
where $\Z_\Bc=[\z(\x_i)]_{\x_i\in\Bc}$ is the representations of the mini-batch $\Bc$.
Note that the dimension of $\Z_\Bc$ is significantly lower than that of $\btheta_{sh}$,
and the left-hand-side $\|\sum_{l=1}^K\tau_l\nabla_{\btheta_{sh}}J_l\|\to0$ if $\|\sum_{l=1}^K\tau_l\nabla_{\Z_\Bc}J_l\|\to 0$.
Hence, we indirectly optimize \eqref{eq:d-npsvcpp-tau} by minimizing its upper bound: 
\begin{equation}
    \btau^\star=\arg\min_{\btau\in\Delta}\frac12\left\|\sum_{l=1}^K\tau_l\nabla_{\Z_\Bc}J_l\right\|^2-\gamma\sum_{l=1}^K\tau_lJ_l.\label{eq:tau-d-npsvc++}
\end{equation}
Additionally, the optimal solution $\btau^\star$ is probably sparse.
The sparse $\btau^\star$ potentially hinders the gradient propagation for the objectives corresponding to the zero entries,
and further causes slow convergence.
To alleviate this problem,
we adopt a momentum strategy to update $\btau$:
\begin{equation}
    \btau_{t+1}=\beta\btau^\star + (1-\beta)\btau_t, \label{eq:tau-momentum}
\end{equation}
where $0<\beta<1$ is a momentum hyperparameter.
We set $\beta=0.85$ in the experiments.
The details of training D-NPSVC++ are stated by Algorithm \ref{alg:D-NPSVCPP} in Appendix \ref{app:alg}.


%% file: exp.tex
\section{Experiments}
In the experiments, we first compared the proposed K-NPSVC++ with both the generally used and the latest NPSVCs.
Next, we evaluate D-NPSVC++ in comparison to the DNNs equipped with the learning criteria.
For fairness, the same network architecture is used for these methods.
Finally, the behaviors of both K-NPSVC++ and D-NPSVC++ are studied empirically. 
More implementation details are shown in Appendix \ref{app:alg}.
The extra experimental results are also provided in Appendix \ref{app:exp}.



\begin{table}[t]
    \renewcommand\arraystretch{1.05}
    \centering
    \caption{Classification performance of different deep models (the first four rows) and kernel machines (the last two rows).
    The kernel machines use the features of the pretrained ResNet34 (without fine-tuning). We highlight \topone{the best} and \topsec{the second best} results.}
    \vspace{-2pt}
    \scalebox{0.84}{
    \begin{tabular}{c|ccc}
    \hline\hline
    \multirow{2}*{Training Criterion} & \multicolumn{3}{c}{Datasets} \\
     \cline{2-4}
    & Cifar-10 & DTD & FLW-102 \\
    \hline
    Cross-entropy & 95.19 & 64.47 & 81.67\\
    \cite{deng2019arcface} & \topsec{95.33} & 64.47 &  82.45\\
    \cite{Li2022DeepTWSVM} & 95.30 & 65.11 &  \topsec{84.44}\\
    D-NPSVC++ (ours) & \topone{96.77}& \topsec{65.90} &  \topone{86.27}\\
    \hline
    TWSVM & 89.83 & 64.94 & 77.70\\
    K-NPSVC++ (ours) & 91.13& \topone{66.54} & 80.63\\
    \hline\hline
    \end{tabular}
    }
    \label{tab:exp-deep}
    \vspace{-10pt}
\end{table}
\subsection{Comparison with SVMs}
We compared K-NPSVC++ with the existing SVMs, particularly NPSVCs,
including vanilla SVM \cite{cortes1995svm},
TWSVM \cite{Shao2011TBSVM}, NPSVM \cite{Shao2014NPSVM}, pinTWSVM \cite{Xu2016pinTWSVM} and RMTBSVM \cite{Zhang2023RMTBSVM}.
The datasets include the CHG dataset for hand gesture recognition \cite{Kim2009CHG},
BinAlpha for hand-written recognition\footnote{https://cs.nyu.edu/\texttildelow roweis/data/}, and datasets from LIBSVM website\footnote{https://www.csie.ntu.edu.tw/\texttildelow cjlin/libsvmtools/datasets/}.
The average accuracy of 10 runs with random split, where the 60\% samples of the whole dataset are used for training and the rest for testing, is reported in Table \ref{tab:exp-svm}.
It can be seen that K-NPSVC++ always outperforms the other compared methods, except for the DNA dataset.
Moreover, we conducted paired t-tests to evaluate the statistical significance.
The results show that the performance of K-NPSVC++ and other methods has no significant difference in the DNA dataset.
Nevertheless, K-NPSVC++ still significantly performs better in most cases,
and it shows significance against all methods on CHG and USPS datasets.
These demonstrate the superiority of K-NPSVC++.

\subsection{Real-world Image Classification}\label{sec:exp-img}
D-NPSVC++ was compared DNNs with the criteria including cross-entropy (i.e, softmax regression), margin softmax regression in \cite{Wang2018MarginSoftmax}, and deep TWSVM \cite{Li2022DeepTWSVM}.
Three image classification datasets are involved, i.e.,
Cifar-10 \cite{cifar10}, DTD \cite{dtd}, and Flowers-102 (FLW-102 for short) \cite{flowers102} datasets.
We also include the comparison with the kernel machines, TWSVM and K-NPSVC++, with the feature extracted by ResNet34 as their input.
As shown in Table \ref{tab:exp-deep}, D-NPSVC++ has beaten other criteria on three datasets.
Compared with deep TWSVM \cite{Li2022DeepTWSVM},
D-NPSVC++ shows a significant improvement in classification performance,
which validates the effectiveness of the proposed training criterion of NPSVC++.
Interestingly,
K-NPSVC++ defeats DNNs and achieves the best performance on the DTD dataset.
The reason may be that K-NPSVC++ found a better feature space in RKHS than DNNs on the DTD dataset,
This phenomenon shows that K-NPSVC++, as a kernel machine, still has competitive power against D-NPSVC++.
Another advantage of K-NPSVC++ is that it needs significantly less training time and computational power than D-NPSVC++ (see the extra results in Appendix \ref{app:exp}).
However, on the other hand, K-NPSVC++ also needs extra memory to store the kernel matrices, which might be the bottleneck in large-scale classification.
Therefore, we conclude that both K- and D-NPSVC++ present promising performance,
and it is also pivotal to select either K- or D-NPSVC++ for classification according to the requirements of performance and computation resources.

\subsection{Empirical Studies of NPSVC++}
\begin{figure}[t]
    \centering
    \subfigure[K-NPSVC++ on BinAlpha]{
        \includegraphics[width=0.222\textwidth]{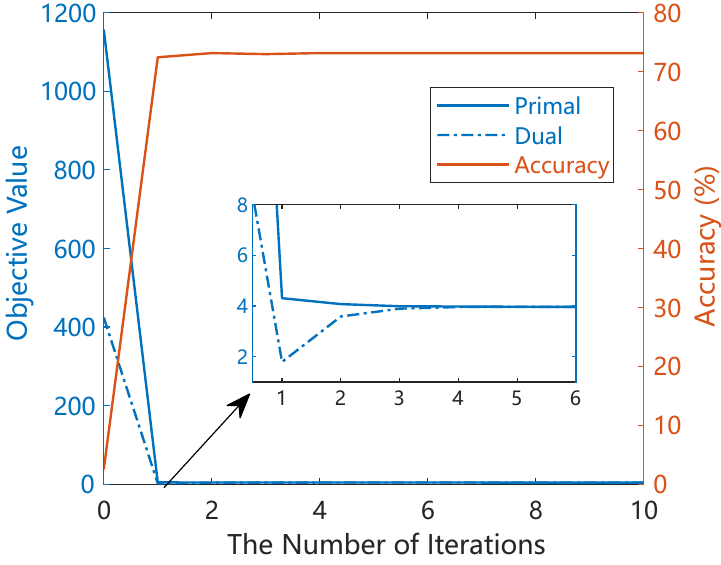}
        \label{fig:conv-knpsvcpp}
    }
    \subfigure[D-NPSVC++ on FLW-102]{
        \includegraphics[width=0.222\textwidth]{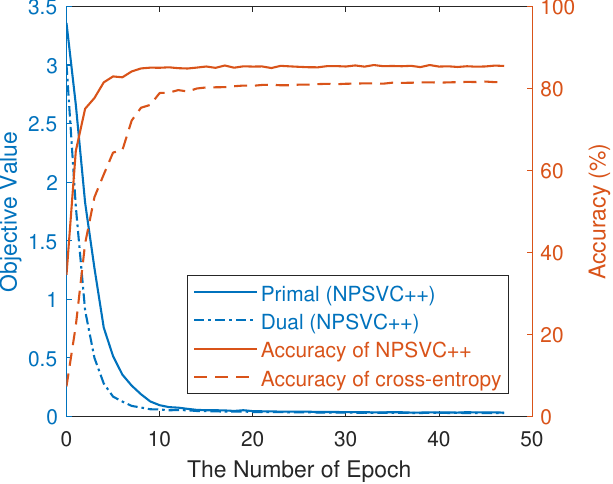}
        \label{fig:conv-dnpsvcpp}
    }
    \vspace{-8pt}
    \caption{Training evolution of K- and D-NPSVC++.}
    \vspace{-12pt}
\end{figure}
\begin{figure}[t]
    \centering
    \subfigure[$\bphi(\x)$]{
        \includegraphics[width=0.22\textwidth]{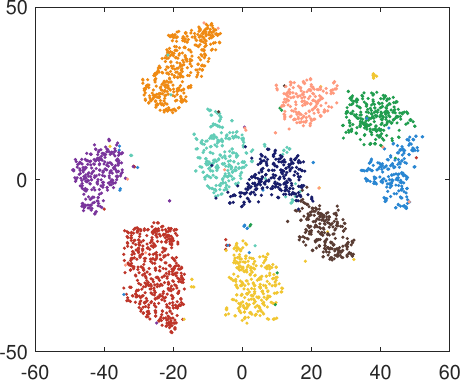}
    }
    \subfigure[$\z(\x)$]{
        \includegraphics[width=0.22\textwidth]{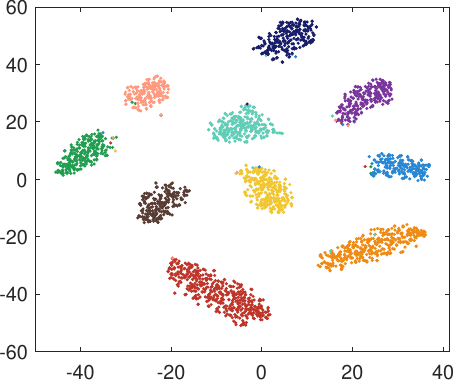}
    }
    \vspace{-8pt}
    \caption{TSNE visualization of raw data and learned features.}
    \vspace{-12pt}
    \label{fig:vis}
\end{figure}
\textbf{Convergence.}
We depict the convergence of K-NPSVC++ in Figure \ref{fig:conv-knpsvcpp}.
The lines ``primal'' and ``dual'' refer to the objective of problem \eqref{eq:Wcheb-prob} and \eqref{eq:dual-Wcheb-prob}.
Intuitively, it can be seen that the primal objective decreases rapidly and merges with the dual objective within a few steps, which indicates optimality since the dual gap becomes zero.
Meanwhile, the accuracy also grows fast and becomes stable finally.
Empirically, K-NPSVC++ can reach convergence within 5 iterations.
Figure \ref{fig:conv-dnpsvcpp} shows the training evolution of D-NPSVC++.
Similar to K-NPSVC++, the primal and dual objectives of D-NPSVC++ decrease monotonically and converge to a minimum.
Compared with the cross-entropy,
D-NPSVC++ achieves the highest accuracy within fewer epochs,
which suggests that D-NPSVC++ could perform more efficient convergence than the softmax regression.

\textbf{Learned features.}
The learned features of K-NPSVC++ are illustrated in Figure \ref{fig:vis}.
It can be observed that the learned features $\z(\x)$ are more discriminative than the raw features $\bphi(\x)$.
The intra-class distance is squeezed and the inter-class margin is enlarged.
This suggests that K-NPSVC++ can learn discriminative features and effectively improve data separability, thus further enhancing performance.

\begin{figure}[t]
    \centering
    \includegraphics[width=0.44\textwidth]{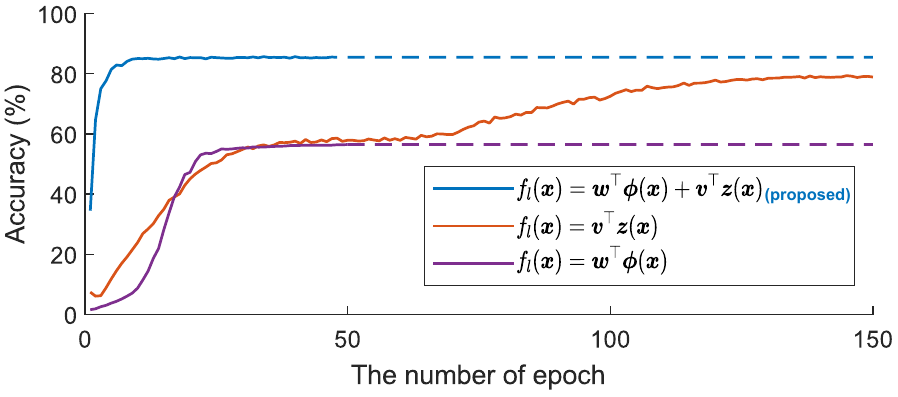}
    \vspace{-10pt}
    \caption{Ablation study of D-NPSVC++ in hypothesis function.
    Dashed lines indicate the convergence accuracy.}
    \vspace{-12pt}
    \label{fig:ablation}
\end{figure}

\textbf{Hypothesis function.}
We conducted an ablation study to examine the rationality of the proposed hypothesis \eqref{eq:npsvcpp-f},
which is compared with D-NPSVC++ trained with $f_l(\x)=\vbf_l^\T\z(\x)$ and $f_l(\x)=\w_l^\T\bphi(\x)$, i.e., taking only one part in \eqref{eq:npsvcpp-f} as the compared hypothesis.
As shown in Figure \ref{fig:ablation}, the proposed hypothesis significantly outperforms the compared two hypotheses.
Intuitively, $f_l(\x)=\w_l^\T\bphi(\x)$ performs the worst as it only depends on the features of the prior encoder.
In contrast, $f_l(\x)=\vbf_l^\T\z(\x)$ achieves a better accuracy yet converges slowly.
Its reason is that D-NPSVC++ should update $\btau$ by minimizing $\|\sum_{l=1}^K\tau_l\nabla_{\Z_{\Bc}}J_l\|$, so the gradient is shrunk and the training is slowed consequently.
Fortunately, the skip connection structure of the proposed hypothesis $f_l(\x)=\w_l^\T\bphi(\x)+\vbf_l^\T\z(\x)$ can effectively avoid this phenomenon.
These results sufficiently validate the efficacy of the designed hypothesis function.

%% file: concl.tex
\section{Conclusion and Future Directions}
In this paper, we establish a new classification framework, NPSVC++,
which is the first end-to-end method integrating representation learning and the nonparallel classifier based on multi-objective optimization.
NPSVC++ aims to achieve the Pareto optimality of multiple dependent objectives,
and it theoretically ensures the optimality of the learned features across classes,
solving the issues of feature suboptimality and class dependency naturally.
We propose the iterative optimization framework and its two realizations, K- and D-NPSVC++,
whose effectiveness is verified in experiments.
NPSVCs are powerful yet underutilized classifiers in machine learning.
They were born with distinctive designs compared to the mainstream classifiers.
Although many works have been proposed,
the inherent mechanism of NPSVC has not been well studied so far.
We believe that NPSVC++ paves a new path and provides insights to investigate these classifiers.
Diverse multi-objective optimization methods could be the future work,
and the generalization ability of NPSVC++ is also an appealing direction.

%% file: appendix.tex
\renewcommand{\theequation}{A.\arabic{equation}}
\renewcommand{\thefigure}{A.\arabic{figure}}
\renewcommand{\thetable}{A.\arabic{table}}
\setcounter{equation}{0}
\setcounter{figure}{0}
\setcounter{table}{0}

\section{Proof of Theorem \ref{thm:mul-obj}}\label{app:proof-thm-mul-obj}
\begin{proof}
    We first rewrite weighted Chebyshev problem as
    \begin{equation}
        \min_{\rho,\btheta}~\rho,\quad\sto\bm J(\btheta)\leq\rho\onef.
    \end{equation}
    Then its Lagrangte function is
    \begin{equation}
        L(\rho,\btheta,\btau)=\rho+\btau^\T(\bm J(\btheta)-\rho\onef).
    \end{equation}
    By KKT conditions, we have
    \begin{align*}
        \nabla_{\rho}L&=1-\btau^\T\onef=0\\
        \nabla_{\btheta}L&=\nabla_\btheta\bm J\btau=\sum_{l=1}^K\tau_l\nabla_{\btheta}J_l=\bm0,\quad \tau_l\geq 0,\forall l\in[K].
    \end{align*}
    where $\nabla_\btheta\bm J=[\nabla_{\btheta}J_1,\nabla_{\btheta}J_2,\cdots,\nabla_{\btheta}J_K]$.
    Therefore, the dual problem is formulated as
    \begin{align*}
        &\mathrel{\phantom{=}}\max_{\btau}\min_{\btheta,\rho}L(\rho,\btheta,\btau)\\
        &=\max_{\btau}\min_{\btheta,\rho}\rho(1-\btau^\T\onef)+\btau^\T\bm J(\btheta)\\
        &=\max_{\btau}\min_{\btheta}\btau^\T\bm J(\btheta),\\
        &\sto\sum_{l=1}^K\tau_l\nabla_{\btheta}J_l=\bm0,~\tau_l\geq 0,~\btau^\T\onef=1.
    \end{align*}
    Besides, let $\btheta_l$ be the specific variable of objective $J_l$, and $\btheta_{sh}$ be the variable shared to all objectives. 
    By the optimality condition of minimization problem $\min_{\btheta}\btau^\T\bm J(\btheta)$, we have
    \begin{equation*}
        \nabla_{\btheta_l}J_l(\btheta)=\bm0,~\forall l\in[K].\label{eq:app:theta-grad-zero}
    \end{equation*}
    \begin{align*}
        \sum_{l=1}^K\tau_l\nabla_{\btheta}J_l
        &=\begin{bmatrix}
            \nabla_{\btheta_1}J_1&\nabla_{\btheta_2}J_2&\cdots&\nabla_{\btheta_K}J_K\\
            \nabla_{\btheta_{sh}}J_1&\nabla_{\btheta_{sh}}J_2&\cdots&\nabla_{\btheta_{sh}}J_K
        \end{bmatrix}\btau\\
        &=\begin{bmatrix}
            \bm0&\bm0&\cdots&\bm0\\
            \nabla_{\btheta_{sh}}J_1&\nabla_{\btheta_{sh}}J_2&\cdots&\nabla_{\btheta_{sh}}J_K
        \end{bmatrix}\btau
        =\begin{bmatrix}
            \bm0\\\bm0
        \end{bmatrix},
    \end{align*}
    we only need to guarantee the constraints:
    \begin{equation*}
        \sum_{l=1}^K\tau_l\nabla_{\btheta_{sh}}J_l=\bm0.
    \end{equation*}
    Thus, the desired result is obtained.    
\end{proof}

\section{Derivation of $\ubf$-step update.}\label{app:u}
$\ubf_l$ \textbf{step}.
By introducing slack variable $\bxi_l$,
we obtain the following problem that optimizes $\ubf_l$:
\begin{align}
    \min_{\bbeta_l,\bxi_l}&~\frac12\|\K_{l}^\T\bbeta\|^2+c\onef^\T\bxi_l+\frac{r_1}{2}(\bbeta_l^\T\K\bbeta_l-2\bbeta_l^\T\K\A\vbf_l)\notag\\
    \sto&\onef-\bxi_l-\K_{-l}^\T\bbeta_l\leq\bm0,\quad\bxi_l\geq\bm 0,\label{eq:NPSVCPlus-u-prob}
\end{align}
where $\K_{-l}=[\bm\kappa_i]_{y_i\not=l}$.
To cope with the inequality constraints,
the Lagrange function is introduced:
\begin{align}
    L(\ubf_l,\bxi_l,\blambda,\bnu)
    &=\frac12\|\K_{l}^\T\bbeta\|^2+c\onef^\T\bxi_l+\frac{r_1}{2}(\bbeta_l^\T\K\bbeta_l-2\bbeta_l^\T\K\A\vbf_l)\notag\\
    &+\blambda^\T(\onef-\bxi_l-\K_{-l}^\T\bbeta_l)-\bnu^\T\bxi,
\end{align}
where $\blambda$ and $\bnu$ are non-negative Lagrange multipliers.
According to Karush-Kuhn-Tucker (KKT) optimality conditions,
\begin{align*}
    \nabla_{\ubf_l}L&=\K_{l}\K_{l}^\T\ubf_l+r_1(\K\bbeta_l-\K\A\vbf_l)-\K_{-l}\blambda=\bm0\\
    \nabla_{\bxi}L&=c\onef-\blambda-\bnu=\bm0.
\end{align*}
Thus the closed-form solution to $\ubf_l$ is given by:
\begin{equation}
    \bbeta_l=(\K_{l}\K_{l}^\T+r_1\K)^{-1}(r_1\K\A\vbf_l-\K_{-l}\blambda_l).\label{eq:kahftw-beta}
\end{equation}
Plugging the above results into the Lagrange function and with some simple algebra,
we derive the dual problem of the primal problem \eqref{eq:NPSVCPlus-u-prob}:
\begin{align}
               &\max_{\blambda,\bnu}\min_{\ubf_l,\bxi}L(\ubf_l,\bxi,\blambda,\bnu)\quad\sto \lambda_i\geq0,\nu_i\geq 0\notag\\
\Leftrightarrow&\min_{\blambda}\frac12\blambda_l^\T\K_{-l}^\T\M_l\K_{-l}\blambda_l+\left(r_1\K_{l}^\T\M_{l}\K\A\vbf_l-\onef\right)^\T\blambda_l,\notag\\
               &~\sto 0\leq\lambda_i\leq c.\label{eq:ahftw-bn-dual}    
\end{align}
where $\M_l=(\K_{l}\K_{l}^\T+r_1\K)^{-1}$.

Note that the computation of $\M_l$ requires matrix inversion and thus costs $O(n^3)$ time.
To simplify the computation,
we first introduce three auxiliary matrices:
\begin{equation}
\begin{aligned}
    \Q^{(l)}_{K,K}&=\K_{-l}\M_l\K_{-l}^\T, \quad\Q^{(l)}_{\Psi,\Psi}=\bPsi^\T\M_l\bPsi,\\
    \Q^{(l)}_{K,\Psi}&=\bPsi^\T\M_l\K_{-l}.
\end{aligned}
\label{eq:kNPSVCPlus-Qmat}
\end{equation}
Then by \eqref{eq:kahftw-beta} and the definition of $\wh\U$, we have
\begin{equation}
    \wh\ubf_l=r_1\Q^{(l)}_{\Psi,\Psi}\wh\Pbf\vbf_l+\Q^{(l)}_{K,\Psi}\blambda_l,\label{eq:kNPSVCPlus-U}
\end{equation}
where $\blambda\in\Rb^{n_l}_+$ denotes the Lagrange multiplier and is obtained by the following dual problem:
\begin{equation}
    \min_{\bm0\leq\blambda_l\leq c\onef}\frac12\blambda_l^\T\Q^{(l)}_{K,K}\blambda_l+(\Q^{(l)}_{K,\Psi}\wh\Pbf\vbf_l-\onef)^\T\blambda_l.\label{eq:kNPSVCPlus-U-qp}
\end{equation}
Now we show how to compute the frequently used matrices \eqref{eq:kNPSVCPlus-Qmat}.
Using the Woodbury formula \cite{Ke2020Welsch}, we rewrite $\M_l$ as
\begin{align*}
    &\mathrel{\phantom{=}}(\K_{l}\K_{l}^\T+r_1\K)^{-1}\\
    &=\frac{1}{r_1}\K^{-1}-\frac{1}{r_1^2}\K^{-1}\K_{l}\left(\Ibf+\frac{1}{r_1}\K_{l}^\T\K^{-1}\K_{l}\right)^{-1}\K_{l}^\T\K^{-1}\\
    &=\frac{1}{r_1}\left[\K^{-1}-\K^{-1}\K_{l}\left(r_1\Ibf+\K_{l}^\T\K^{-1}\K_{l}\right)^{-1}\K_{l}^\T\K^{-1}\right].
\end{align*}
Remind the Cholesky decomposition $\K=\bPsi\bPsi^\T$,
and thus $\K^{-1}=\bPsi^{-\T}\bPsi^{-1}$. 
Then we have
\begin{equation}
    \K_{l}^\T\K^{-1}\K_{l}=\bPsi_{l}\bPsi^\T(\bPsi^{-\T}\bPsi^{-1})\bPsi\bPsi_{l}^\T=\bPsi_{l}\bPsi_{l}^\T,
\end{equation}
where $\bPsi_l\in\Rb^{n_l\times n}$ is comprised of the rows of $\bPsi$ corresponding to the $l$-th class.
Likewise, we have $\K_{-l}^\T\K^{-1}\K_{-l}=\bPsi_{-l}\bPsi^\T_{-l}$.
Therefore, the following equations hold:
\begin{equation}
\begin{aligned}
    \Q^{(l)}_{\Psi,\Psi}&=\frac{1}{r_1}\left[\Ibf-\bPsi_{l}^\T(r_1\Ibf+\bPsi_{l}\bPsi_{l}^\T)^{-1}\bPsi_{l}\right],\\
    \Q^{(l)}_{K,\Psi}&=\bPsi_{-l}^\T\Q^{(l)}_{\Psi,\Psi},\quad \Q^{(l)}_{K,K}=\bPsi_{-l}^\T\Q^{(l)}_{\Psi,\Psi}\bPsi_{-l},
\end{aligned}
\end{equation}
Therefore, the size of matrix inversion is decreased to $n_l$,
which will significantly reduce the computation burden.

\section{Details of the Algorithms and Implementations} \label{app:alg}
\subsection{Algorithms}
The detailed training procedures are shown in Algorithm \ref{alg:K-NPSVCPP} and \ref{alg:D-NPSVCPP}.
\begin{algorithm}[t]
    \caption{Training K-NPSVC++}
    \label{alg:K-NPSVCPP}
    \textbf{Input}: Samples $\X\in\Rb^{m\times n}, \y\in[K]^{n}$.\\
    \textbf{Output}: Matrix $\wh\U\in\Rb^{n\times K}$, $\V\in\Rb^{d\times K}$, and projection matrix $\wh\Pbf\in\Rb^{n\times d}$.
    \begin{algorithmic}[1] 
        \STATE Initialize $\wh\U_0,\V_0,\wh\Pbf_0$ such that $\wh\Pbf_0^\T\wh\Pbf_0=\Ibf$, and $\btau_0=\onef/K$.
        \STATE Let $t\leftarrow 0$.
        \IF{``Linear kernel'' is used}
            \STATE Let $\bPsi\leftarrow\X^\T$
        \ELSE
            \STATE Compute the kernel matrix $\K=\Kc(\X,\X)+\epsilon\Ibf$ and
            \STATE Compute the Cholesky decomposition $\bPsi\bPsi^\T=\K$.
        \ENDIF
        \STATE Compute $\Hbf\leftarrow\mu\bPsi^\T\Lbf\bPsi$.
        \STATE Compute $\Q^{(l)}_{\Psi,\Psi}$, $\Q^{(l)}_{\Psi,K}$ and $\Q^{(l)}_{K,K}$ for $l\in[K]$ via \eqref{eq:kNPSVCPlus-Qmat-compute}.
        \WHILE{$t<\text{MAX\_ITER}$ and not convergent}
            \FOR{$l\in[K]$}
                \STATE Solve the problem \eqref{eq:kNPSVCPlus-U-qp}.
                \STATE Update $\wh\ubf_{t+1,l}\leftarrow r_1\Q^{(l)}_{\Psi,\Psi}\wh\Pbf_t\vbf_{t,l}+\Q^{(l)}_{K,\Psi}\blambda_l$.
            \ENDFOR
            \STATE Update $\V_{t+1}\leftarrow r_1\wh\Pbf_t^\T\wh\U_{t+1}/(r_1+r_2)$.
            \STATE Update $\E\leftarrow r_1\wh\U_{t+1}\diag(\btau_t)\V_{t+1}^\T$.
            \STATE Update $\wh\Pbf_{t+\frac12}\leftarrow\mathsf{GPI}(\Hbf,\E,\Pbf_{t})$ with Algorithm by \cite{nie2017GPI}.
            \STATE Update $\btau_{t+1}$ by solving QPP \eqref{eq:NPSVCPlus-tau}.
            \STATE Let $\wh\Pbf_{t+1}\leftarrow\bpi_{\St}\left(\wh\Pbf_{t+\frac12}-\eta\sum_{l=1}^K\tau_{t+1,l}\nabla_{\wh\Pbf_{t+\frac12}}^\St J_l\right)$.
            \STATE Let $t\leftarrow t+1$.
        \ENDWHILE
    \end{algorithmic}
\end{algorithm}

\begin{algorithm}[h]
    \caption{Training D-NPSVC++}
    \label{alg:D-NPSVCPP}
    \begin{algorithmic}[1] 
        \FOR{$p=1:n_{epoch}$}
            \FOR{each mini-batch $\Bc=\{\x_{i_1},\cdots,\x_{i_{|\Bc|}}\}$}
            \STATE Activate the parameter $\tilde\W$.
            \STATE \verb+# First-step + \eqref{eq:theta-step}.
            \STATE Cache the representations from prior encoder $\bPhi_\Bc\leftarrow [\bphi(\x_{i_1}),\cdots,\bphi(\x_{i_{|\Bc|}})]$.
            \STATE Compute $\Z_\Bc\leftarrow [\z(\bphi(\x_{i_1})),\cdots,\z(\bphi(\x_{i_{|\Bc|}}))]$, and let $\tilde\Z_{\Bc}=\mathrm{Concat}(\bPhi_{\Bc},\Z_{\Bc})$.
            \STATE Compute $\hat f_{il}\leftarrow \tilde\w_l^\T\tilde\z_{i},~\forall i\in[i_1,\cdots,i_{\Bc}],~l\in[K]$.
            \STATE Compute the objectives for $\forall l\in[K]$:
            $
                J_l=\frac{1}{|\Bc|}\left[\sum_{y_i=l}\ell_{s}(\hat f_{il})+\frac{c}{K-1}\sum_{y_i\not=l}\ell_{d}(\hat f_{il})\right].$
            \STATE Compute $J=\sum_{l=1}^K\tau_lJ_l$, and apply gradient back-propagation and update parameters $[\tilde\W,\btheta_{sh}]$.
            \STATE \verb+# Second-step + \eqref{eq:tau-step}.
            \STATE Freeze the parameter $\tilde\W$.
            \STATE Compute $\Z_\Bc\leftarrow [\z(\bphi(\x_{i_1})),\cdots,\z(\bphi(\x_{|\Bc|}))]$, and let $\tilde\Z_{\Bc}=\mathrm{Concat}(\bPhi_{\Bc},\Z_{\Bc})$.
            \STATE Again compute the objectives $J_l$ and the gradient $\nabla_{\Z_{\Bc}}J_l$ for $\forall l\in[K]$.
            \STATE Solve the QPP \eqref{eq:tau-d-npsvc++} and update $\btau$ with \eqref{eq:tau-momentum}.
            \STATE Compute $J=\sum_{l=1}^K\tau_lJ_l$, and apply gradient back-propagation and update the shared parameters $\btheta_{sh}$.
            \ENDFOR
        \ENDFOR
    \end{algorithmic}
\end{algorithm}
\subsection{Implementations}
The code of K- and D-NPSVC++ is available at \url{https://github.com/Apple-Zhang/NPSVCPP}.
\subsubsection{K-NPSVC++}
In our experiments, we use Gaussian kernel $\Kc(\x,\x')=\exp(-\|\x-\x'\|^2/t)$ and $t$ is set as the mean of pair-wise squared distance: $t=\sum_{i,j}\|\x-\x'\|^2/n^2$. This kernel setting is also applied to other SVMs.
The graph $\G$ is constructed as a $k$ nearest neighbor graph following \cite{he2005LPP}, i.e,
\begin{equation}
    G_{ij}=\begin{cases}
        \Kc(\x_i,\x_j),& i\in\Nc_k(j)\vee j\in\Nc_k(i)\\
        0,& \text{Otherwise.}
    \end{cases}
\end{equation}
where $\Kc(\cdot,\cdot)$ is the Gaussian kernel defined above, and $\Nc_k(i)$ denotes the index set of the $k$ nearest neighborhood of $\x_i$.
The hyperparameter $k$, for convenience, is set as $k=\lfloor\log_2n\rfloor$,
where $n$ is the number of the training samples.
In practical implementation,
we use the normalized Laplacian matrix $\Lbf=\Ibf-\D^{-1/2}\G\D^{-1/2}$ for better stability.
K-NPSVC++ is implemented with MATLAB R2023a.

\subsubsection{D-NPSVC++}
\begin{figure}[h]
    \centering
    \includegraphics[width=0.98\textwidth]{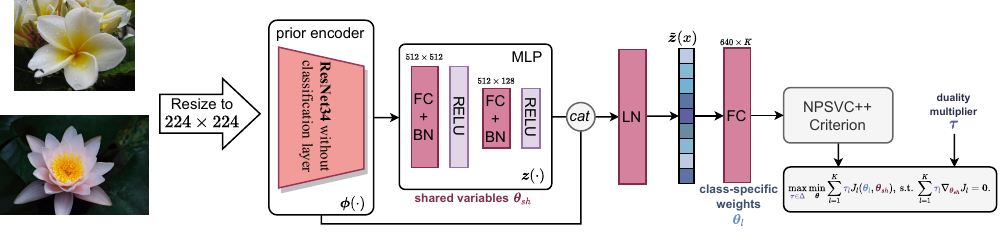}
    \caption{The architecture of D-NPSVC++ in the experiments. The prior encoder in the experiments is ResNet34. The ``FC'', ``BN'', and ``LN'' denote fully connected, batch normalization, and layer normalization layers respectively.}
    \label{fig:app:arch}
\end{figure}

The implementation of D-NPSVC++ is based on PyTorch\footnote{https://pytorch.org/}.
In our experiments, we apply the square loss as similarity loss and squared-hinge loss as dissimilarity loss, i.e,
\begin{equation}
    \ell_s(f_l(\x_i))=\frac12[f_l(\x_i)]^2,\quad \ell_d(f_l(\x_i))=\frac12[s-f_l(x_i)]^2_+.
\end{equation}
Different from K-NPSVC++, whose dissimilarity loss is non-squared hinge loss, i.e., $\ell_d(f_l(\x_i))=[s-f_l(x_i)]_+$, D-NPSVC++ uses squared-hinge loss due to its differentiability.
In the experiments, we utilize the architectures of Figure \ref{fig:app:arch} for D-NPSVC++.
The compared methods also apply the same architecture with D-NPSVC++.

We train the model for 50 epochs and use the batch size 64.
The prior encoder is fine-tuned in the first 10 epochs and frozen in the remained epochs.
The model is trained with stochastic gradient descent (i.e., SGD) with momentum, 
where the momentum parameter is 0.9.
The weight decay is set as $10^{-4}$.
The initial learning rate is set as $10^{-3}$ for Cifar-10 and DTD datasets,
and $10^{-2}$ for FLW-102 dataset.
We use the cosine annealing strategy proposed by \cite{loshchilov2016sgdr} to schedule the learning rate, with the maximal iteration 50 and the minimal learning rate $10^{-5}$.
We dynamically adjust hyperparameter $\gamma$ during training via $\gamma=\frac{\gamma_{\max{}}}{2}[1+\cos(p\pi/T_{\max{}})]+\epsilon$, where $p$ denotes the number of epoch and $\epsilon$ is a small number.
These settings are also applied to the compared methods.

\section{Extra Experimental Results}\label{app:exp}
\subsection{Training time}
We add the results of training time of the experiments in Section \ref{sec:exp-img}.
\begin{table}[t]
    \renewcommand\arraystretch{1.1}
    \centering
    \caption{Training time of different DNNs (the first four rows) and kernel machines (the last two rows). The experiments are run on the workstation with CPU: Intel Xeon Silver 4110 @ 2.10GHz, GPU: NVIDIA RTX 2080Ti.}
    \scalebox{0.92}{
    \begin{tabular}{c|c|ccc}
    \hline\hline
    \multirow{2}*{Model type} & \multirow{2}*{Training Criterion} & \multicolumn{3}{c}{Datasets} \\
     \cline{3-5}
    && Cifar-10 & DTD & FLW-102 \\
    \hline
    \multirow{4}*{DNNs (with GPU)}&Softmax Regression& 169min & 41min & 72min\\
    &\cite{deng2019arcface} & 169min & 43min  & 71min \\
    &\cite{Li2022DeepTWSVM} & 124min & 42min &  82min\\
    &D-NPSVC++ (ours) & 218min & 41min &  96min\\
    \hline
    \multirow{2}*{Kernel machines (with CPU)} & TWSVM & 33min & 21.69s & 12.67s\\
    &K-NPSVC++ (ours) & 56min& 34.31s & 20.24s\\
    \hline\hline
    \end{tabular}
    }
    \label{tab:exp-time}                              
\end{table}

\subsection{Hyperparameter of K-NPSVC++}
\begin{figure}[h]
    \centering
    \includegraphics[width=0.6\textwidth]{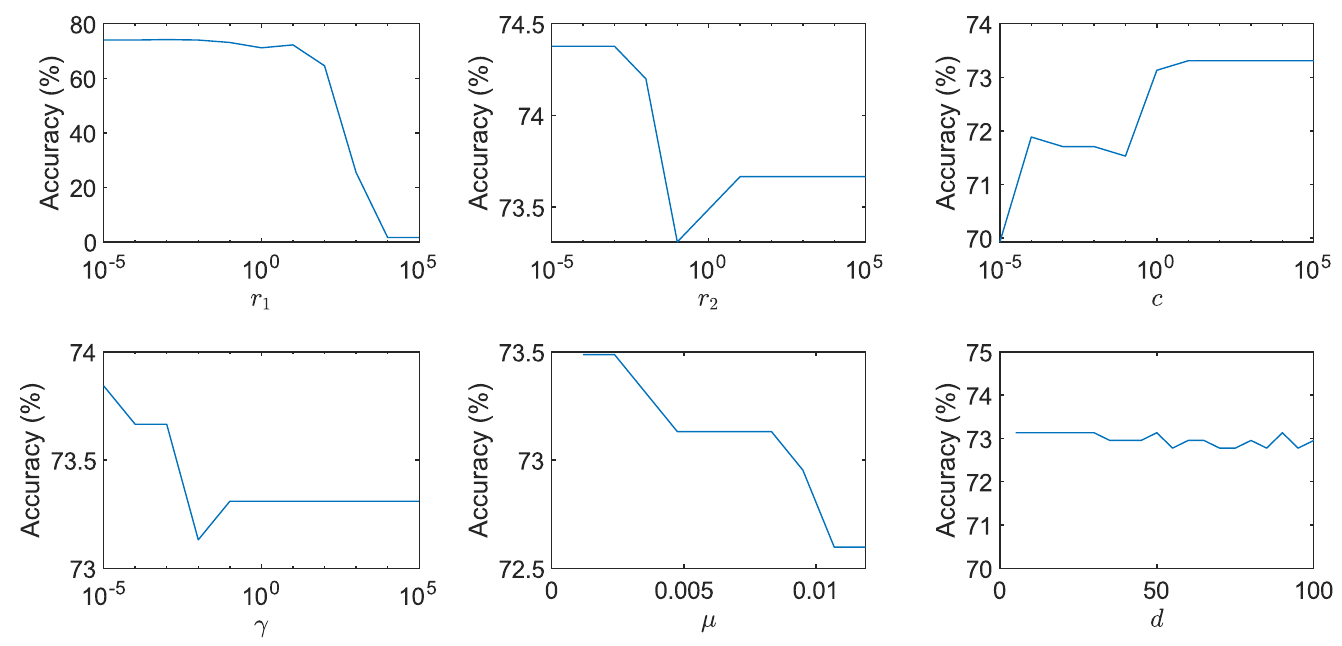}
    \caption{The evaluation of hyperparameter on BinAlpha dataset.}
    \label{fig:hyperpara-binalpha}
\end{figure}

\begin{figure}[h]
    \centering
    \includegraphics[width=0.6\textwidth]{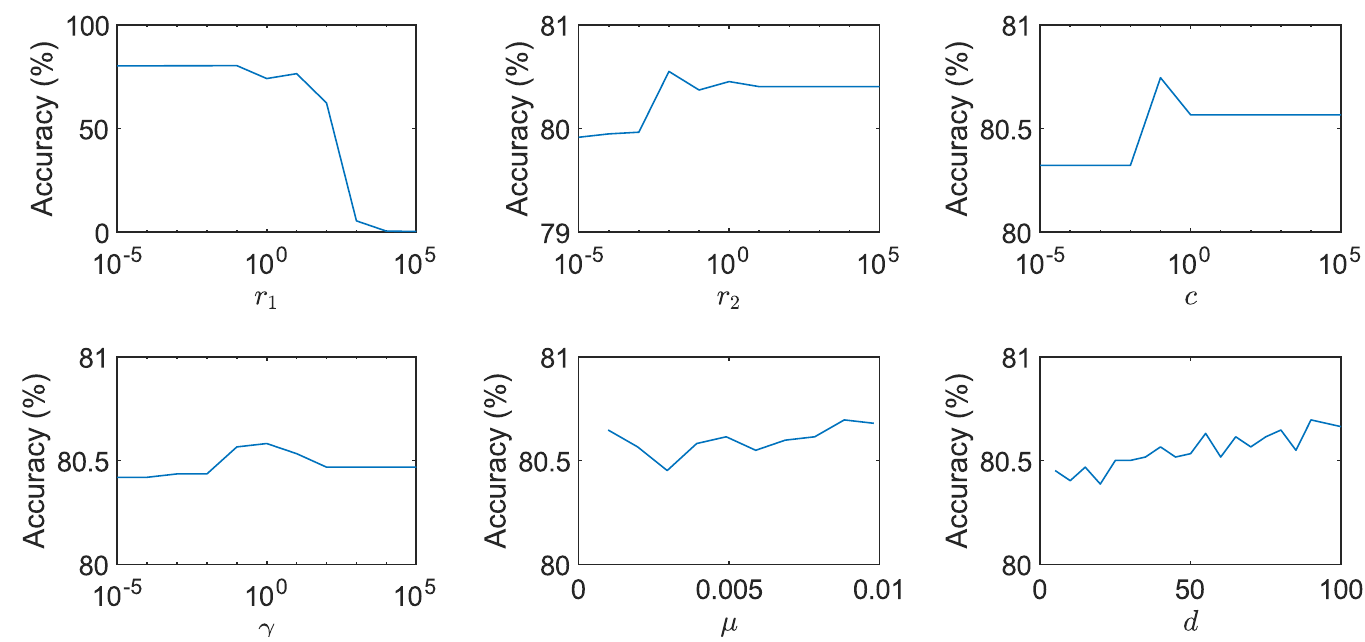}
    \caption{The evaluation of hyperparameter on FLW-102 dataset (features extracted by ResNet34).}
    \label{fig:hyperpara-flw102}
\end{figure}

Since K-NPSVC++ has a lot of hyperparameters,
we should study its performance versus different hyperparameter settings.
The experimental results are shown in Figure \ref{fig:hyperpara-binalpha} and \ref{fig:hyperpara-flw102}.
We observed that K-NPSVC++ works stably for small $r_1$,
but performs poorly when $r_1$ is extremely large.
When $r_1$ is large, the model tends to make $\|\w_l\|\to0$,
or equivalently, $\ubf_l-\Pbf\vbf_l\to\bm0$,
which probably leads to a trivial subspace and results in model degradation.
Therefore, a moderately small $r_1$ is preferred for K-NPSVC++.
Nevertheless, K-NPSVC++ is relatively robust to other hyperparameters. 


\subsection{Objectives Visualization}
\begin{figure}[t]
    \centering
    \subfigure[The first epoch]{
        \includegraphics[width=0.31\textwidth]{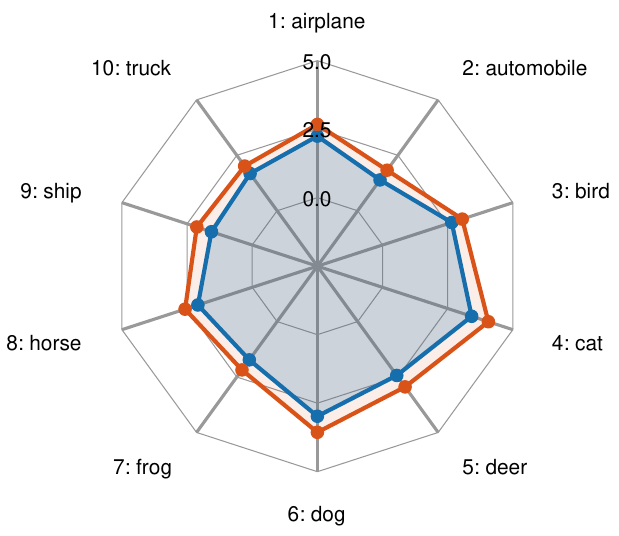}
    }
    \subfigure[The 25-th epoch (log-scaled axes)]{
        \includegraphics[width=0.31\textwidth]{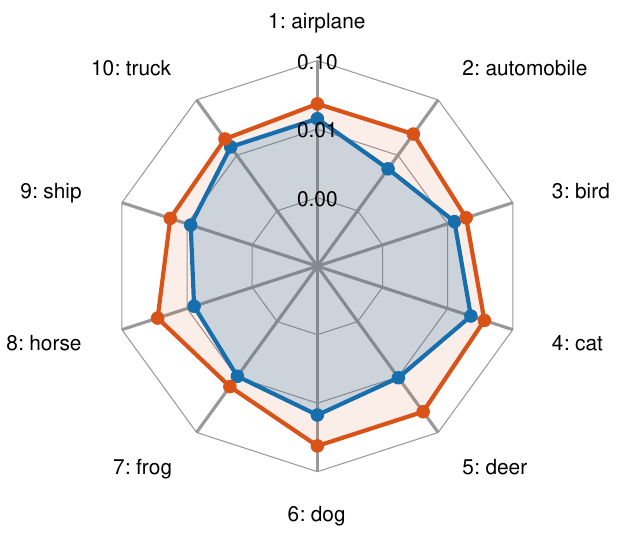}
    }
    \subfigure[The final epoch (log-scaled axes)]{
        \includegraphics[width=0.31\textwidth]{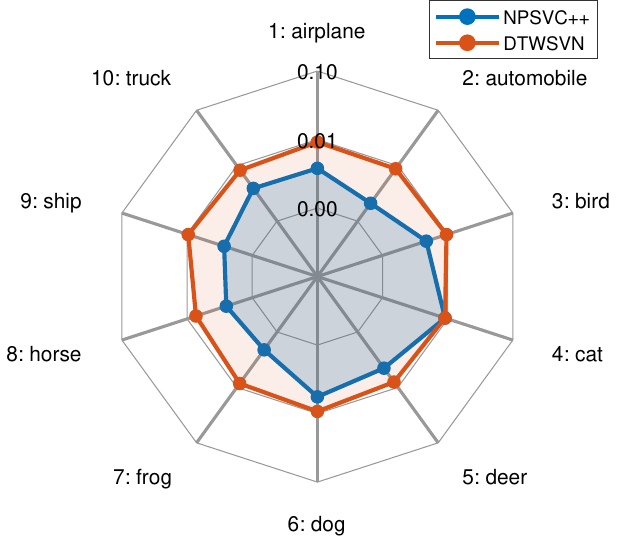}
    }
    \caption{Radar chart of the objectives of D-NPSVC++ and DTWSVN on Cifar-10.
    We plot the objective of each class in the first epoch, the 25-th epoch, and the final epoch. 
    }
    \label{fig:radar}
\end{figure}
To show that the proposed D-NPSVC++ achieves a better solution to the multi-objective optimization problem \eqref{eq:NPSVC-mul-obj} than DTWSVN \cite{Li2022DeepTWSVM},
we plot the radar charts to illustrate their objectives of each class in Figure \ref{fig:radar}.
Both two methods compute the objective of $l$-th class by
\begin{equation}
    J_l=\frac{1}{n}\left[\sum_{y_i=l}\ell_{s}(f_l(\x_i))+\frac{c}{K-1}\sum_{y_i\not=l}\ell_{d}(f_l(\x_i))\right],
\end{equation}
where the two methods use the same loss functions.
It can be observed that D-NPSVC++ always achieve lower objectives in all classes than DTWSVN,
which suggests the supority of D-NPSVC++ over DTWSVN and the effectiveness of the proposed training strategy.
To explain this phenomenon, we compare the optimization criterion of these two methods:
\begin{align}
    \text{DTWSVN: }&\min_{\btheta} \frac1K\sum_{l=1}^K J_l,\\
    \text{D-NPSVC++: }&\begin{cases}
        \btheta\text{ step:}&\min_{\btheta}\sum_{l=1}^K\tau_lJ_l\\
        \btau\text{ step:}&\max_{\btau\in\Delta}\sum_{l=1}^K\tau_lJ_l,~\sto\sum_{l=1}^K\tau_l\nabla_{\Z_{\Bc}}J_l=\bm0,
    \end{cases}
\end{align}
DTWSVN uses uniform scaling methods to minimize multiple dependent objectives, and usually fails to achieve the optimal solution \cite{Multitask2018Sener}.
The reason is that it does not consider the relationship between different objectives.
In contrast, the proposed D-NPSVC++ aims to achieve Pareto optimality,
which implements the trade-off of different objectives.
Thus, D-NPSVC++ can obtain a better solution than DTWSVN.


%% file: main.bbl
\begin{thebibliography}{49}
\providecommand{\natexlab}[1]{#1}
\providecommand{\url}[1]{\texttt{#1}}
\expandafter\ifx\csname urlstyle\endcsname\relax
  \providecommand{\doi}[1]{doi: #1}\else
  \providecommand{\doi}{doi: \begingroup \urlstyle{rm}\Url}\fi

\bibitem[{Arun Kumar} \& Gopal(2009){Arun Kumar} and Gopal]{ArunKumar2009LSTWSVM}
{Arun Kumar}, M. and Gopal, M.
\newblock Least squares twin support vector machines for pattern classification.
\newblock \emph{Expert Systems with Applications}, 36\penalty0 (4):\penalty0 7535--7543, 2009.

\bibitem[Bai et~al.(2014)Bai, Wang, Shao, and Deng]{Bai2014FeaSelTWSVM}
Bai, L., Wang, Z., Shao, Y.-H., and Deng, N.-Y.
\newblock A novel feature selection method for twin support vector machine.
\newblock \emph{Knowledge-Based Systems}, 59:\penalty0 1--8, 2014.

\bibitem[Beck(2014)]{beck2014DoubleCoordinate}
Beck, A.
\newblock The 2-coordinate descent method for solving double-sided simplex constrained minimization problems.
\newblock \emph{Journal of Optimization Theory and Applications}, 162:\penalty0 892--919, 2014.

\bibitem[Boyd et~al.(2004)Boyd, Boyd, and Vandenberghe]{boyd2004CVXOPT}
Boyd, S., Boyd, S.~P., and Vandenberghe, L.
\newblock \emph{Convex optimization}.
\newblock {Cambridge University press}, 2004.

\bibitem[Chandra \& Bedi(2021)Chandra and Bedi]{chandra2021survey}
Chandra, M.~A. and Bedi, S.
\newblock Survey on {SVM} and their application in image classification.
\newblock \emph{International Journal of Information Technology}, 13:\penalty0 1--11, 2021.

\bibitem[Cimpoi et~al.(2014)Cimpoi, Maji, Kokkinos, Mohamed, and Vedaldi]{dtd}
Cimpoi, M., Maji, S., Kokkinos, I., Mohamed, S., and Vedaldi, A.
\newblock Describing textures in the wild.
\newblock In \emph{Proceedings of the IEEE Conference on Computer Vision and Pattern Recognition}, 2014.

\bibitem[Cortes \& Vapnik(1995)Cortes and Vapnik]{cortes1995svm}
Cortes, C. and Vapnik, V.
\newblock Support-vector networks.
\newblock \emph{Machine learning}, 20:\penalty0 273--297, 1995.

\bibitem[Deng et~al.(2019)Deng, Guo, Xue, and Zafeiriou]{deng2019arcface}
Deng, J., Guo, J., Xue, N., and Zafeiriou, S.
\newblock Arcface: Additive angular margin loss for deep face recognition.
\newblock In \emph{Proceedings of the IEEE/CVF conference on computer vision and pattern recognition}, pp.\  4690--4699, 2019.

\bibitem[Diaz-Vico et~al.(2020)Diaz-Vico, Prada, Omari, and Dorronsoro]{diaz2020deepSVM}
Diaz-Vico, D., Prada, J., Omari, A., and Dorronsoro, J.
\newblock Deep support vector neural networks.
\newblock \emph{Integrated Computer-Aided Engineering}, 27\penalty0 (4):\penalty0 389--402, 2020.

\bibitem[Dosovitskiy et~al.(2020)Dosovitskiy, Beyer, Kolesnikov, Weissenborn, Zhai, Unterthiner, Dehghani, Minderer, Heigold, Gelly, et~al.]{dosovitskiy2020ViT}
Dosovitskiy, A., Beyer, L., Kolesnikov, A., Weissenborn, D., Zhai, X., Unterthiner, T., Dehghani, M., Minderer, M., Heigold, G., Gelly, S., et~al.
\newblock An image is worth 16x16 words: Transformers for image recognition at scale.
\newblock In \emph{International Conference on Learning Representations}, 2020.

\bibitem[Désidéri(2012)]{MGDA2012}
Désidéri, J.-A.
\newblock Multiple-gradient descent algorithm ({MGDA}) for multiobjective optimization.
\newblock \emph{Comptes Rendus Mathematique}, 350\penalty0 (5-6):\penalty0 313--318, 2012.

\bibitem[Gao et~al.(2011)Gao, Ye, and Ye]{Gao2021L1TWSVM}
Gao, S., Ye, Q., and Ye, N.
\newblock 1-norm least squares twin support vector machines.
\newblock \emph{Neurocomputing}, 74\penalty0 (17):\penalty0 3590--3597, 2011.

\bibitem[Giagkiozis \& Fleming(2015)Giagkiozis and Fleming]{Giagkiozis2015MultiObj}
Giagkiozis, I. and Fleming, P.
\newblock Methods for multi-objective optimization: An analysis.
\newblock \emph{Information Sciences}, 293:\penalty0 338--350, 2015.
\newblock ISSN 0020-0255.

\bibitem[Goudjil et~al.(2018)Goudjil, Koudil, Bedda, and Ghoggali]{goudjil2018novel}
Goudjil, M., Koudil, M., Bedda, M., and Ghoggali, N.
\newblock A novel active learning method using {SVM} for text classification.
\newblock \emph{International Journal of Automation and Computing}, 15:\penalty0 290--298, 2018.

\bibitem[He et~al.(2016)He, Zhang, Ren, and Sun]{He2016ResNet}
He, K., Zhang, X., Ren, S., and Sun, J.
\newblock Deep residual learning for image recognition.
\newblock In \emph{IEEE/CVF Conference on Computer Vision and Pattern Recognition}, pp.\  770--778, 2016.

\bibitem[He et~al.(2005)He, Yan, Hu, Niyogi, and Zhang]{he2005LPP}
He, X., Yan, S., Hu, Y., Niyogi, P., and Zhang, H.-J.
\newblock Face recognition using laplacianfaces.
\newblock \emph{IEEE Transactions on Pattern Analysis and Machine Intelligence}, 27\penalty0 (3):\penalty0 328--340, 2005.

\bibitem[Jayadeva et~al.(2007)Jayadeva, Khemchandani, and Chandra]{Jayadeva2007twsvm}
Jayadeva, Khemchandani, R., and Chandra, S.
\newblock Twin support vector machines for pattern classification.
\newblock \emph{IEEE Transactions on Pattern Analysis and Machine Intelligence}, 29\penalty0 (5):\penalty0 905--910, 2007.

\bibitem[Ke et~al.(2022)Ke, Gong, Liu, Zhao, Yang, and Tao]{Ke2020Welsch}
Ke, J., Gong, C., Liu, T., Zhao, L., Yang, J., and Tao, D.
\newblock Laplacian welsch regularization for robust semisupervised learning.
\newblock \emph{IEEE Transactions on Cybernetics}, 52\penalty0 (1):\penalty0 164--177, 2022.

\bibitem[Kim \& Cipolla(2009)Kim and Cipolla]{Kim2009CHG}
Kim, T.-K. and Cipolla, R.
\newblock Canonical correlation analysis of video volume tensors for action categorization and detection.
\newblock \emph{IEEE Transactions on Pattern Analysis and Machine Intelligence}, 31\penalty0 (8):\penalty0 1415--1428, 2009.

\bibitem[Krizhevsky(2009)]{cifar10}
Krizhevsky, A.
\newblock Learning multiple layers of features from tiny images.
\newblock Technical report, 2009.

\bibitem[Li \& Yang(2022)Li and Yang]{Li2022DeepTWSVM}
Li, M. and Yang, Z.
\newblock Deep twin support vector networks.
\newblock In \emph{Artificial Intelligence}, pp.\  94--106, 2022.

\bibitem[Liu et~al.(2022)Liu, Mao, Wu, Feichtenhofer, Darrell, and Xie]{liu2022convnet}
Liu, Z., Mao, H., Wu, C.-Y., Feichtenhofer, C., Darrell, T., and Xie, S.
\newblock A convnet for the 2020s.
\newblock In \emph{Proceedings of the IEEE/CVF conference on computer vision and pattern recognition}, pp.\  11976--11986, 2022.

\bibitem[Loshchilov \& Hutter(2016)Loshchilov and Hutter]{loshchilov2016sgdr}
Loshchilov, I. and Hutter, F.
\newblock {SGDR}: Stochastic gradient descent with warm restarts.
\newblock In \emph{International Conference on Learning Representations}, 2016.

\bibitem[Majeed \& Alkhafaji(2023)Majeed and Alkhafaji]{Russel2023AppLSSVM}
Majeed, R.~R. and Alkhafaji, S. K.~D.
\newblock {ECG} classification system based on multi-domain features approach coupled with least square support vector machine ({LS-SVM}).
\newblock \emph{Computer Methods in Biomechanics and Biomedical Engineering}, 26\penalty0 (5):\penalty0 540--547, 2023.

\bibitem[Mangasarian \& Wild(2006)Mangasarian and Wild]{Mangasarian2006GEPSVM}
Mangasarian, O. and Wild, E.
\newblock Multisurface proximal support vector machine classification via generalized eigenvalues.
\newblock \emph{IEEE Transactions on Pattern Analysis and Machine Intelligence}, 28\penalty0 (1):\penalty0 69--74, 2006.

\bibitem[Mao et~al.(2023)Mao, Mohri, and Zhong]{Mao2023CE}
Mao, A., Mohri, M., and Zhong, Y.
\newblock Cross-entropy loss functions: Theoretical analysis and applications.
\newblock In \emph{Proceedings of the 40th International Conference on Machine Learning}, pp.\  23803--23828, 2023.

\bibitem[Mohri et~al.(2018)Mohri, Rostamizadeh, and Talwalkar]{mohri2018FoML}
Mohri, M., Rostamizadeh, A., and Talwalkar, A.
\newblock \emph{Foundations of machine learning}.
\newblock MIT press, 2018.

\bibitem[Momma et~al.(2022)Momma, Dong, and Liu]{momma2022multiobj}
Momma, M., Dong, C., and Liu, J.
\newblock A multi-objective/multi-task learning framework induced by {Pareto} stationarity.
\newblock In \emph{International Conference on Machine Learning}, pp.\  15895--15907, 2022.

\bibitem[Nie et~al.(2017)Nie, Zhang, and Li]{nie2017GPI}
Nie, F., Zhang, R., and Li, X.
\newblock A generalized power iteration method for solving quadratic problem on the stiefel manifold.
\newblock \emph{Science China Information Sciences}, 60:\penalty0 273--297, 2017.

\bibitem[Nilsback \& Zisserman(2008)Nilsback and Zisserman]{flowers102}
Nilsback, M.-E. and Zisserman, A.
\newblock Automated flower classification over a large number of classes.
\newblock In \emph{Indian Conference on Computer Vision, Graphics and Image Processing}, Dec 2008.

\bibitem[Sch{\"o}lkopf et~al.(2002)Sch{\"o}lkopf, Smola, Bach, et~al.]{scholkopf2002kernel}
Sch{\"o}lkopf, B., Smola, A.~J., Bach, F., et~al.
\newblock \emph{Learning with kernels: support vector machines, regularization, optimization, and beyond}.
\newblock MIT press, 2002.

\bibitem[Sener \& Koltun(2018)Sener and Koltun]{Multitask2018Sener}
Sener, O. and Koltun, V.
\newblock Multi-task learning as multi-objective optimization.
\newblock In \emph{Advances in Neural Information Processing Systems}, 2018.

\bibitem[Shao et~al.(2011)Shao, Zhang, Wang, and Deng]{Shao2011TBSVM}
Shao, Y.-H., Zhang, C.-H., Wang, X.-B., and Deng, N.-Y.
\newblock Improvements on twin support vector machines.
\newblock \emph{IEEE Transactions on Neural Networks}, 22\penalty0 (6):\penalty0 962--968, 2011.

\bibitem[Shao et~al.(2014)Shao, Chen, and Deng]{Shao2014NPSVM}
Shao, Y.-H., Chen, W.-J., and Deng, N.-Y.
\newblock Nonparallel hyperplane support vector machine for binary classification problems.
\newblock \emph{Information Sciences}, 263:\penalty0 22--35, 2014.

\bibitem[Simonyan \& Zisserman(2015)Simonyan and Zisserman]{Simonyan2015VGG}
Simonyan, K. and Zisserman, A.
\newblock Very deep convolutional networks for large-scale image recognition.
\newblock In \emph{International Conference on Learning Representations}, pp.\  1--14, 2015.

\bibitem[Suykens \& Vandewalle(1999)Suykens and Vandewalle]{suykens1999LSSVM}
Suykens, J.~A. and Vandewalle, J.
\newblock Least squares support vector machine classifiers.
\newblock \emph{Neural Processing Letters}, 9\penalty0 (3):\penalty0 293--300, 1999.

\bibitem[Tanveer et~al.(2019)Tanveer, Sharma, and Suganthan]{TANVEER2019pinTWSVM}
Tanveer, M., Sharma, A., and Suganthan, P.
\newblock General twin support vector machine with pinball loss function.
\newblock \emph{Information Sciences}, 494:\penalty0 311--327, 2019.

\bibitem[Tian et~al.(2014)Tian, Qi, Ju, Shi, and Liu]{Tian2014NPSVM}
Tian, Y., Qi, Z., Ju, X., Shi, Y., and Liu, X.
\newblock Nonparallel support vector machines for pattern classification.
\newblock \emph{IEEE Transactions on Cybernetics}, 44\penalty0 (7):\penalty0 1067--1079, 2014.

\bibitem[Vaswani et~al.(2017)Vaswani, Shazeer, Parmar, Uszkoreit, Jones, Gomez, Kaiser, and Polosukhin]{Vaswani2017Transformer}
Vaswani, A., Shazeer, N., Parmar, N., Uszkoreit, J., Jones, L., Gomez, A.~N., Kaiser, L.~u., and Polosukhin, I.
\newblock Attention is all you need.
\newblock In \emph{Advances in Neural Information Processing Systems}, 2017.

\bibitem[Wang et~al.(2018)Wang, Cheng, Liu, and Liu]{Wang2018MarginSoftmax}
Wang, F., Cheng, J., Liu, W., and Liu, H.
\newblock Additive margin softmax for face verification.
\newblock \emph{IEEE Signal Processing Letters}, 25\penalty0 (7):\penalty0 926--930, 2018.

\bibitem[Wang et~al.(2019)Wang, Feng, and Wu]{wang2019svm}
Wang, J., Feng, K., and Wu, J.
\newblock {SVM}-based deep stacking networks.
\newblock In \emph{Proceedings of the AAAI conference on artificial intelligence}, volume~33, pp.\  5273--5280, 2019.

\bibitem[Wen \& Yin(2013)Wen and Yin]{wen2013Stiefel}
Wen, Z. and Yin, W.
\newblock A feasible method for optimization with orthogonality constraints.
\newblock \emph{Mathematical Programming}, 142\penalty0 (1):\penalty0 397--434, 2013.

\bibitem[Xie et~al.(2020)Xie, Nie, and Gao]{xie2020mmpp}
Xie, D., Nie, F., and Gao, Q.
\newblock On the optimal solution to maximum margin projection pursuit.
\newblock \emph{Multimedia Tools and Applications}, 79:\penalty0 35441--35461, 2020.

\bibitem[Xie et~al.(2013)Xie, Hone, Xie, Gao, Shi, and Liu]{Xie2013OVRTWSVM}
Xie, J., Hone, K., Xie, W., Gao, X., Shi, Y., and Liu, X.
\newblock Extending twin support vector machine classifier for multi-category classification problems.
\newblock \emph{Intelligent Data Analysis}, 17\penalty0 (4):\penalty0 649--664, 2013.

\bibitem[Xie et~al.(2023)Xie, Li, and Sun]{Xie2023DeepMultiview}
Xie, X., Li, Y., and Sun, S.
\newblock Deep multi-view multiclass twin support vector machines.
\newblock \emph{Information Fusion}, 91:\penalty0 80--92, 2023.

\bibitem[Xu et~al.(2017)Xu, Yang, and Pan]{Xu2016pinTWSVM}
Xu, Y., Yang, Z., and Pan, X.
\newblock A novel twin support-vector machine with pinball loss.
\newblock \emph{IEEE Transactions on Neural Networks and Learning Systems}, 28\penalty0 (2):\penalty0 359--370, 2017.

\bibitem[Yuan et~al.(2021)Yuan, Yang, and Sun]{Yuan2021CTWSVM}
Yuan, C., Yang, L., and Sun, P.
\newblock Correntropy-based metric for robust twin support vector machine.
\newblock \emph{Information Sciences}, 545:\penalty0 82--101, 2021.

\bibitem[Zhang et~al.(2023)Zhang, Lai, Kong, and Shen]{Zhang2023RMTBSVM}
Zhang, J., Lai, Z., Kong, H., and Shen, L.
\newblock Robust twin bounded support vector classifier with manifold regularization.
\newblock \emph{IEEE Transactions on Cybernetics}, 53\penalty0 (8):\penalty0 5135--5150, 2023.

\bibitem[Zou et~al.(2006)Zou, Hastie, and Tibshirani]{zou2006spca}
Zou, H., Hastie, T., and Tibshirani, R.
\newblock Sparse principal component analysis.
\newblock \emph{Journal of computational and graphical statistics}, pp.\  265--286, 2006.

\end{thebibliography}
